\documentclass[english,letterpaper]{article}
\usepackage[T1]{fontenc}
\usepackage[latin9]{inputenc}
\usepackage{array}
\usepackage{float}
\usepackage{mathtools}
\usepackage{multirow}
\usepackage{amsmath}
\usepackage{amsthm}
\usepackage{amssymb}
\usepackage{graphicx}

\makeatletter

\newcommand{\noun}[1]{\textsc{#1}}
\providecommand{\tabularnewline}{\\}
\floatstyle{ruled}
\newfloat{algorithm}{tbp}{loa}
\providecommand{\algorithmname}{Algorithm}
\floatname{algorithm}{\protect\algorithmname}

 \usepackage{times}
 \usepackage{graphicx} 

 \usepackage{natbib}

 \usepackage{algorithm}
 \usepackage{algorithmic}

 \usepackage{hyperref}

 \usepackage[accepted]{icml2016}

\theoremstyle{plain}
\newtheorem{thm}{\protect\theoremname}
  \theoremstyle{remark}
  \newtheorem{claim}[thm]{\protect\claimname}

\@ifundefined{showcaptionsetup}{}{%
 \PassOptionsToPackage{caption=false}{subfig}}
\usepackage{subfig}
\makeatother

\usepackage{babel}
  \providecommand{\claimname}{Claim}
\providecommand{\theoremname}{Theorem}

\begin{document}
\twocolumn[
\icmltitle{Network Morphism}

\icmlauthor{Tao Wei$^\dagger$}{taowei@buffalo.edu}
\icmlauthor{Changhu Wang}{chw@microsoft.com}
\icmlauthor{Yong Rui}{yongrui@microsoft.com}
\icmlauthor{Chang Wen Chen}{chencw@buffalo.edu}
\icmladdress{Microsoft Research, Beijing, China, 100080.}
\vspace*{-0.12in}
\icmladdress{Department of Computer Science and Engineering, University at Buffalo, Buffalo, NY, 14260.}

\icmlkeywords{boring formatting information, machine learning, ICML}

\vskip 0.3in
]

\begin{abstract}
We present in this paper a systematic study on how to morph a well-trained
neural network to a new one so that its network function can be completely
preserved. We define this as \emph{network morphism} in this research.
After morphing a parent network, the child network is expected to
inherit the knowledge from its parent network and also has the potential
to continue growing into a more powerful one with much shortened training
time. The first requirement for this network morphism is its ability
to handle diverse morphing types of networks, including changes of
depth, width, kernel size, and even subnet. To meet this requirement,
we first introduce the network morphism equations, and then develop
novel morphing algorithms for all these morphing types for both classic
and convolutional neural networks. The second requirement for this
network morphism is its ability to deal with non-linearity in a network.
We propose a family of parametric-activation functions to facilitate
the morphing of any continuous non-linear activation neurons. Experimental
results on benchmark datasets and typical neural networks demonstrate
the effectiveness of the proposed network morphism scheme.
\end{abstract}

\section{Introduction}

\begin{figure}[t]
\begin{centering}
 \includegraphics[width=0.85\linewidth]{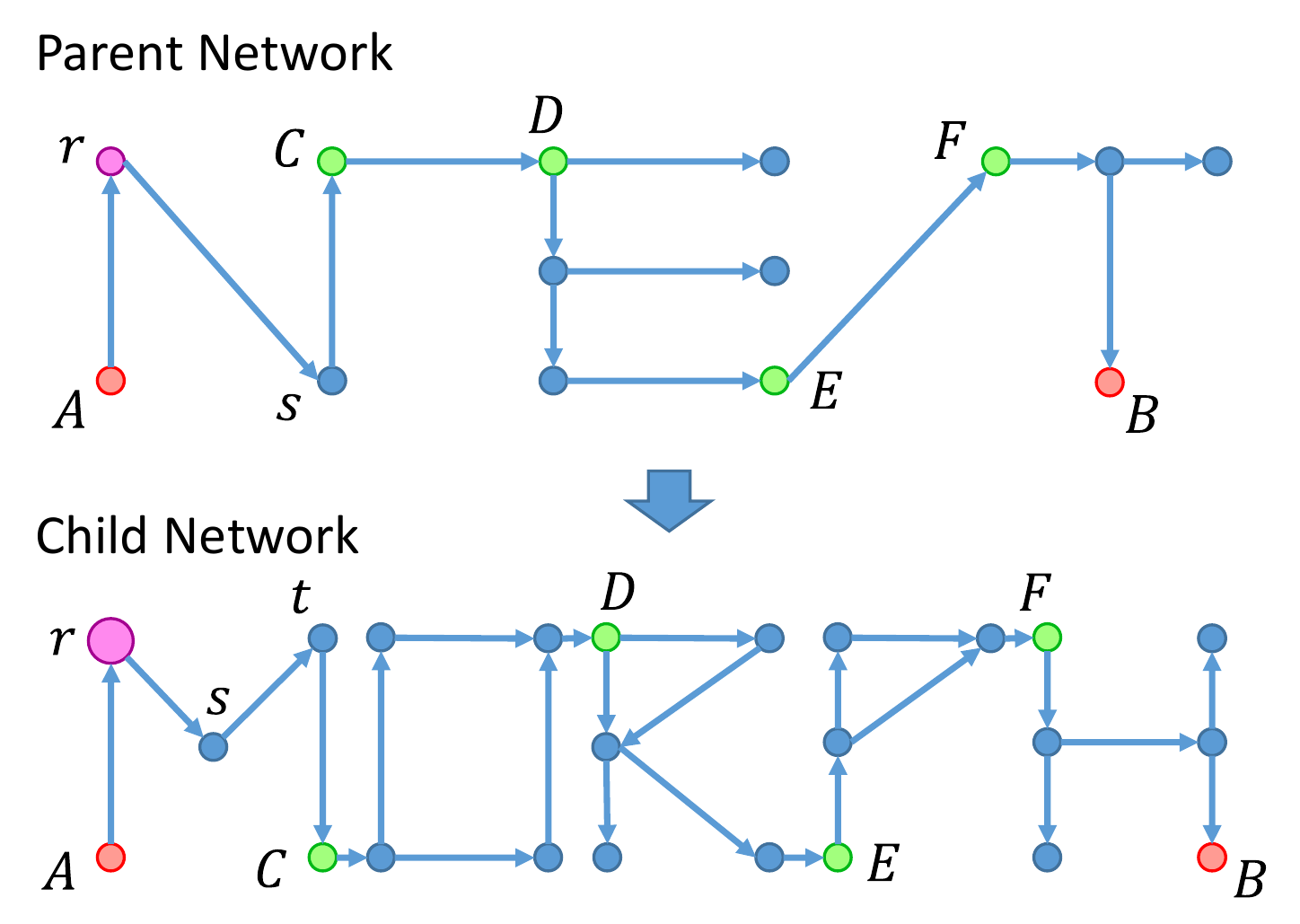}
\par\end{centering}
 \caption{Illustration of network morphism. The child network is expected to
inherit the entire knowledge from the parent network with the network
function preserved. A variety of morphing types are illustrated. The
change of segment $AC$ represents the depth morphing: $s\to s+t$;
the inflated node $r$ involves width and kernel size morphing; a
subnet is embedded in segment $CD$, which is subnet morphing. Complex
network morphism can also be achieved with a combination of these
basic morphing operations. \label{fig:netmorph} }
\end{figure}

Deep convolutional neural networks (DCNNs) have achieved state-of-the-art
results on diverse computer vision tasks such as image classification
\cite{KrizhevskySutskeverHinton2012,SimonyanZisserman2014,SzegedyLiuJiaEtAl2014},
object detection \cite{GirshickDonahueDarrellEtAl2014,Girshick2015,RenHeGirshickEtAl2015},
and semantic segmentation \cite{LongShelhamerDarrell2014}. However,
training such a network is very time-consuming. It usually takes weeks
or even months to train an effective deep network, let alone the exploration
of diverse network settings. It is very much desired for these well-trained
networks to be directly adopted for other related applications with
minimum retraining.

To accomplish such an ideal goal, we need to systematically study
how to morph a well-trained neural network to a new one with its network
function completely preserved. We call such operations network morphism.
Upon completion of such morphism, the child network shall not only
inherit the entire knowledge from the parent network, but also be
capable of growing into a more powerful one in much shortened training
time as the process continues on. This is fundamentally different
from existing work related to network knowledge transferring, which
either tries to mimic a parent network's outputs \cite{BuciluCaruanaNiculescu-Mizil2006,BaCaruana2014,RomeroBallasKahouEtAl2014},
or pre-trains to facilitate the convergence and/or adapt to new datasets
with possible total change in network function \cite{SimonyanZisserman2014,OquabBottouLaptevEtAl2014}.

Mathematically, a morphism is a structure-preserving map from one
mathematical structure to another \cite{Weisstein2002}. In the context
of neural networks, network morphism refers to a parameter-transferring
map from a parent network to a child network that preserves its function
and outputs. Although network morphism generally does not impose constraints
on the architecture of the child network, we limit the investigation
of network morphism to the expanding mode, which intuitively means
that the child network is deeper and/or wider than its parent network.
Fig. \ref{fig:netmorph} illustrates the concept of network morphism,
where a variety of morphing types are demonstrated including depth
morphing, width morphing, kernel size morphing, and subnet morphing.
In this work, we derive network morphism equations for a successful
morphing operation to follow, based on which novel network morphism
algorithms can be developed for all these morphing types. The proposed
algorithms work for both classic multi-layer perceptron models and
convolutional neural networks. Since in the proposed network morphism
it is required that the output is unchanged, a complex morphing can
be decomposed into basic morphing steps, and thus can be solved easily.

Depth morphing is an important morphing type, since current top-notch
neural networks are going deeper and deeper \cite{KrizhevskySutskeverHinton2012,SimonyanZisserman2014,SzegedyLiuJiaEtAl2014,HeZhangRenEtAl2015}.
One heuristic approach is to embed an identity mapping layer into
the parent network, which is referred as IdMorph. IdMorph is explored
by a recent work \cite{ChenGoodfellowShlens2015}, but is potentially
problematic due to the sparsity of the identity layer, and might fail
sometimes \cite{HeZhangRenEtAl2015}. To overcome the issues associated
with IdMorph, we introduce several practices for the morphism operation
to follow, and propose a deconvolution-based algorithm for network
depth morphing. This algorithm is able to asymptotically fill in all
parameters with non-zero elements. In its worst case, the non-zero
occupying rate of the proposed algorithm is still higher than IdMorph
for an order of magnitude.

Another challenge the proposed network morphism will face is the dealing
of the non-linearity in a neural network. Even the simple IdMorph
method fails in this case, because it only works for idempotent functions\footnote{An idempotent function $\varphi$ is defined to satisfy $\varphi\circ\varphi=\varphi$.
This condition passes the ReLU function but fails on most of other
commonly used activation functions, such as Sigmoid and TanH.}. In this work, to deal with the non-linearity, we introduce the concept
of parametric-activation function family, which is defined as an adjoint
function family for arbitrary non-linear activation function. It can
reduce the non-linear operation to a linear one with a parameter that
can be learned. Therefore, the network morphism of any continuous
non-linear activation neurons can be solved.

To the best of our knowledge, this is the first work about network
morphism, except the recent work \cite{ChenGoodfellowShlens2015}
that introduces the IdMorph. We conduct extensive experiments to show
the effectiveness of the proposed network morphism learning scheme
on widely used benchmark datasets for both classic and convolutional
neural networks. The effectiveness of basic morphing operations are
also verified. Furthermore, we show that the proposed network morphism
is able to internally regularize the network, that typically leads
to an improved performance. Finally, we also successfully morph the
well-known 16-layered VGG net \cite{SimonyanZisserman2014} to a better
performing model, with only ${\textstyle \frac{1}{15}}$ of the training
time comparing against the training from scratch.

\section{Related Work}

We briefly introduce recent work related to network morphism and identify
the differences from this work.

\textbf{\emph{Mimic Learning}}. A series of work trying to mimic the
teacher network with a student network have been developed, which
usually need learning from scratch. For example, \cite{BuciluCaruanaNiculescu-Mizil2006}
tried to train a \emph{lighter} network by mimicking an ensemble network.
\cite{BaCaruana2014} extended this idea, and used a \emph{shallower
but wider} network to mimic a deep and wide network. In \cite{RomeroBallasKahouEtAl2014},
the authors adopted a \emph{deeper but narrower} network to mimic
a deep and wide network. The proposed network morphism scheme is different
from these algorithms, since instead of mimicking, its goal is to
make the child network directly inherit the intact knowledge (network
function) from the parent network. This allows network morphism to
achieve the same performance. That is why the networks are called
parent and child, instead of teacher and student. Another major difference
is that the child network is not learned from scratch.

\textbf{\emph{Pre-training and Transfer Learning}}. Pre-training \cite{SimonyanZisserman2014}
is a strategy proposed to facilitate the convergence of very deep
neural networks, and transfer learning\footnote{Although transfer learning in its own concept is very general, here
it is referred as a technique used for DCNNs to pre-train the model
on one dataset and then adapt to another.} \cite{SimonyanZisserman2014,OquabBottouLaptevEtAl2014} is introduced
to overcome the overfitting problem when training large neural networks
on relatively small datasets. They both re-initialize the last few
layers of the parent network with the other layers remaining the same
(or refined in a lighter way). Their difference is that pre-training
continues to train the child network on the same dataset, while transfer
learning continues on a new one. However, these two strategies totally
alter the parameters in the last few layers, as well as the network
function.

\textbf{\emph{Net2Net}}. Net2Net is a recent work proposed in \cite{ChenGoodfellowShlens2015}.
Although it targets at the same problem, there are several major differences
between network morphism and Net2Net. First, the solution of Net2Net
is still restricted to the IdMorph approach, while NetMorph is the
first to make it possible to embed non-identity layers. Second, Net2Net's
operations only work for idempotent activation functions, while NetMorph
is the first to handle arbitrary non-linear activation functions.
Third, Net2Net's discussion is limited to width and depth changes,
while NetMorph studies a variety of morphing types, including depth,
width, kernel size, and subnet changes. Fourth, Net2Net needs to separately
consider depth and width changes, while NetMorph is able to simultaneously
conduct depth, width, and kernel size morphing in a single operation.

\section{Network Morphism}

We shall first discuss the depth morphing in the linear case, which
actually also involves with width and kernel size morphing. Then we
shall describe how to deal with the non-linearities in the neural
networks. Finally, we shall present the stand-alone versions for width
morphing and kernel size morphing, followed by the subnet morphing.

\subsection{Network Morphism: Linear Case}

Let us start from the simplest case of a classic neural network. We
first drop all the non-linear activation functions and consider a
neural network only connected with fully connected layers.

\begin{figure}
\begin{centering}
 \includegraphics[width=0.75\linewidth]{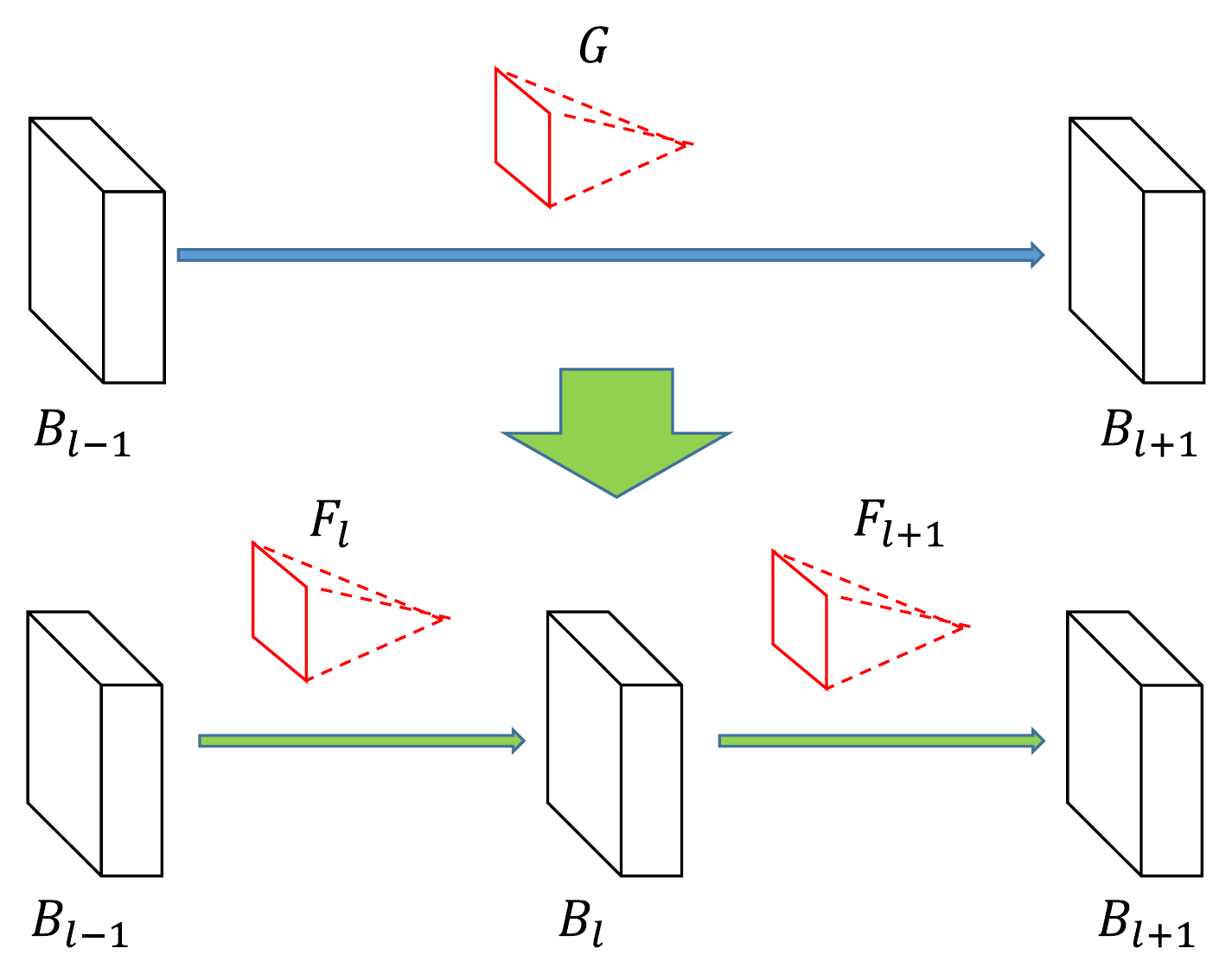}
\par\end{centering}

 \caption{Network morphism linear. $B_{*}$ represents blobs (hidden units),
$G$ and $F_{*}$ are convolutional filters (weight matrices) for
DCNNs (classic neural networks). $G$ is morphed into $F_{l}$ and
$F_{l+1}$, satisfying the network morphism equation (\ref{eq:net_morph}).
\label{fig:net_morph_linear}}

\end{figure}

As shown in Fig. \ref{fig:net_morph_linear}, in the parent network,
two hidden layers $B_{l-1}$ and $B_{l+1}$ are connected via the
weight matrix $G$:
\begin{equation}
B_{l+1}=G\cdot B_{l-1},
\end{equation}
where $B_{l-1}\in\mathbb{R}^{C_{l-1}}$, $B_{l+1}\in\mathbb{R}^{C_{l+1}}$,
$G\in\mathbb{R}^{C_{l+1}\times C_{l-1}}$, $C_{l-1}$ and $C_{l+1}$
are the feature dimensions of $B_{l-1}$ and $B_{l+1}$. For network
morphism, we shall insert a new hidden layer $B_{l}$, so that the
child network satisfies:
\begin{equation}
B_{l+1}=F_{l+1}\cdot B_{l}=F_{l+1}\cdot(F_{l}\cdot B_{l-1})=G\cdot B_{l-1},
\end{equation}
where $B_{l}\in\mathbb{R}^{C_{l}}$, $F_{l}\in\mathbb{R}^{C_{l}\times C_{l-1}}$,
and $F_{l+1}\in\mathbb{R}^{C_{l+1}\times C_{l}}$. It is obvious that
network morphism for classic neural networks is equivalent to a matrix
decomposition problem:
\begin{equation}
G=F_{l+1}\cdot F_{l}.\label{eq:net_morph_classic}
\end{equation}
Next, we consider the case of a deep convolutional neural network
(DCNN). For a DCNN, the build-up blocks are convolutional layers rather
than fully connected layers. Thus, we call the hidden layers as blobs,
and weight matrices as filters. For a 2D DCNN, the blob $B_{*}$ is
a 3D tensor of shape $(C_{*},H_{*},W_{*})$, where $C_{*}$, $H_{*}$,
and $W_{*}$ represent the number of channels, height and width of
$B_{*}$. The filters $G$, $F_{l}$, and $F_{l+1}$ are 4D tensors
of shapes $(C_{l+1},C_{l-1},K,K)$, $(C_{l},C_{l-1},K_{1},K_{1})$,
and $(C_{l+1},C_{l},K_{2},K_{2})$, where $K$, $K_{1}$, $K_{2}$
are convolutional kernel sizes.

The convolutional operation in a DCNN can be defined in a multi-channel
way:
\begin{equation}
B_{l}(c_{l})=\sum_{c_{l-1}}B_{l-1}(c_{l-1})*F_{l}(c_{l},c_{l-1}),
\end{equation}
where $*$ is the convolution operation defined in a traditional way.
It is easy to derive that the filters $F_{l},$ $F_{l+1}$ and $G$
shall satisfy the following equation:
\begin{equation}
\tilde{G}(c_{l+1},c_{l-1})=\sum_{c_{l}}F_{l}(c_{l},c_{l-1})*F_{l+1}(c_{l+1},c_{l}),\label{eq:net_morph_conv}
\end{equation}
where $\tilde{G}$ is a zero-padded version of $G$ whose effective
kernel size (receptive field) is $\tilde{K}=K_{1}+K_{2}-1\geq K$.
If $\tilde{K}=K$, we will have $\tilde{G}=G$.

Mathematically, inner products are equivalent to multi-channel convolutions
with kernel sizes of $1\times1$. Thus, Equation (\ref{eq:net_morph_classic})
is equivalent to Equation (\ref{eq:net_morph_conv}) with $K=K_{1}=K_{2}=1$.
Hence, we are able to unify them into one equation:
\begin{equation}
\tilde{G}=F_{l+1}\circledast F_{l},\label{eq:net_morph}
\end{equation}
where $\circledast$ is a non-communicative operator that can either
be an inner product or a multi-channel convolution. We call Equation
(\ref{eq:net_morph}) as the network morphism equation (for depth
in the linear case).

Although Equation (\ref{eq:net_morph}) is primarily derived for depth
morphing ($G$ morphs into $F_{l}$ and $F_{l+1}$), it also involves
network width $($the choice of $C_{l}$), and kernel sizes (the choice
of $K_{1}$ and $K_{2}$). Thus, it will be called network morphism
equation for short for the remaining of this paper.

The problem of network depth morphing is formally formulated as follows:

\begin{description}
\item [{Input:}] $G$ of shape $(C_{l+1},C_{l-1},K,K)$; $C_{l}$, $K_{1}$,
$K_{2}$. 

\item [{Output:}] $F_{l}$ of shape $(C_{l},C_{l-1},K_{1},K_{1})$, $F_{l+1}$
of shape $(C_{l+1},C_{l},K_{2},K_{2})$ that satisfies Equation (\ref{eq:net_morph}).
\end{description}

\subsection{Network Morphism Algorithms: Linear Case}

In this section, we introduce two algorithms to solve for the network
morphism equation (\ref{eq:net_morph}).

Since the solutions to Equation (\ref{eq:net_morph}) might not be
unique, 
we shall make the morphism operation to follow the desired practices
that: 1) the parameters will contain as many non-zero elements as
possible, and 2) the parameters will need to be in a consistent scale.
These two practices are widely adopted in existing work, since random
initialization instead of zero filling for non-convex optimization
problems is preferred \cite{Bishop2006}, and the scale of the initializations
is critical for the convergence and good performance of deep neural
networks \cite{GlorotBengio2010,HeZhangRenEtAl2015a}.

Next, we introduce two algorithms based on deconvolution to solve
the network morphism equation (\ref{eq:net_morph}), i.e., 1) general
network morphism, and 2) practical network morphism. The former one
fills in all the parameters with non-zero elements under certain condition,
while the latter one does not depend on such a condition but can only
asymptotically fill in all parameters with non-zero elements.

\subsubsection{General Network Morphism}

This algorithm is proposed to solve Equation (\ref{eq:net_morph})
under certain condition. As shown in Algorithm \ref{alg:net_morph_general},
it initializes convolution kernels $F_{l}$ and $F_{l+1}$ of the
child network with random noises. Then we iteratively solve $F_{l+1}$
and $F_{l}$ by fixing the other. For each iteration, $F_{l}$ or
$F_{l+1}$ is solved by deconvolution. Hence the overall loss is always
decreasing and is expected to converge. However, it is not guaranteed
that the loss in Algorithm \ref{alg:net_morph_general} will always
converge to 0.

We claim that if the parameter number of either $F_{l}$ or $F_{l+1}$
is no less than $\tilde{G}$, Algorithm \ref{alg:net_morph_general}
shall converge to 0.
\begin{claim}
\label{claim:net_morph_converge}If the following condition is satisfied,
the loss in Algorithm \ref{alg:net_morph_general} shall converge
to 0 (in one step): {\small{}
\begin{equation}
\min(C_{l}C_{l-1}K_{1}^{2},C_{l+1}C_{l}K_{2}^{2})\geq C_{l+1}C_{l-1}(K_{1}+K_{2}-1)^{2}.\label{eq:net_morph_converge}
\end{equation}
} The three items in condition (\ref{eq:net_morph_converge}) are
the parameter numbers of $F_{l}$ ,$F_{l+1}$, and $\tilde{G}$, respectively.
\end{claim}
It is easy to check the correctness of Condition (\ref{eq:net_morph_converge}),
as a multi-channel convolution can be written as the multiplication
of two matrices. Condition (\ref{eq:net_morph_converge}) claims that
we have more unknowns than constraints, and hence it is an undetermined
linear system. Since random matrices are rarely inconsistent (with
probability 0), the solutions of the undetermined linear system always
exist.

\subsubsection{Practical Network Morphism}

\begin{algorithm}[tb]
\protect\caption{General Network Morphism}
\label{alg:net_morph_general} \begin{algorithmic}

\STATE \textbf{Input:} $G$ of shape $(C_{l+1},C_{l-1},K,K)$; $C_{l}$,
$K_{1}$, $K_{2}$

\STATE \textbf{Output: }$F_{l}$ of shape $(C_{l},C_{l-1},K_{1},K_{1})$,
$F_{l+1}$ of shape $(C_{l+1},C_{l},K_{2},K_{2})$

\STATE Initialize $F_{l}$ and $F_{l+1}$ with random noise.

\STATE Expand $G$ to $\tilde{G}$ with kernel size $\tilde{K}=K_{1}+K_{2}-1$
by padding zeros.

\REPEAT

\STATE Fix $F_{l}$, and calculate $F_{l+1}=deconv(\tilde{G},F_{l})$

\STATE Fix $F_{l+1}$, and calculate $F_{l}=deconv(\tilde{G},F_{l+1})$

\STATE Calculate loss $l=\|\tilde{G}-conv(F_{l},F_{l+1})\|^{2}$

\UNTIL{$l=0$ or $maxIter$ is reached}

\STATE Normalize $F_{l}$ and $F_{l+1}$ with equal standard variance.

\end{algorithmic}
\end{algorithm}

\begin{algorithm}[tb]
\protect\caption{Practical Network Morphism}
\label{alg:net_morph_smart} \begin{algorithmic}

\STATE \textbf{Input:} $G$ of shape $(C_{l+1},C_{l-1},K,K)$; $C_{l}$,
$K_{1}$, $K_{2}$

\STATE \textbf{Output: }$F_{l}$ of shape $(C_{l},C_{l-1},K_{1},K_{1})$,
$F_{l+1}$ of shape $(C_{l+1},C_{l},K_{2},K_{2})$

\STATE /{*} For simplicity, we illustrate this algorithm for the
case `$F_{l}$ expands $G$' {*}/

\STATE $K_{2}^{r}=K_{2}$

\REPEAT

\STATE Run Algorithm \ref{alg:net_morph_general} with $maxIter$
set to 1: $l,F_{l},F_{l+1}^{r}=\textsc{NetMorphGeneral}(G;C_{l},K_{1},K_{2}^{r})$

\STATE $K_{2}^{r}=K_{2}^{r}-1$

\UNTIL{$l=0$}

\STATE Expand $F_{l+1}^{r}$ to $F_{l+1}$ with kernel size $K_{2}$
by padding zeros.

\STATE Normalize $F_{l}$ and $F_{l+1}$ with equal standard variance.

\end{algorithmic}
\end{algorithm}

Next, we propose a variant of Algorithm \ref{alg:net_morph_general}
that can solve Equation (\ref{eq:net_morph}) with a sacrifice in
the non-sparse practice. This algorithm reduces the zero-converging
condition to that the parameter number of either $F_{l}$ or $F_{l+1}$
is no less than $G$, instead of $\tilde{G}$. Since we focus on network
morphism in an expanding mode, we can assume that this condition is
self-justified, namely, either $F_{l}$ expands $G$, or $F_{l+1}$
expands $G$. Thus, we can claim that this algorithm solves the network
morphism equation (\ref{eq:net_morph}). As described in Algorithm
\ref{alg:net_morph_smart}, for the case that $F_{l}$ expands $G$,
starting from $K_{2}^{r}=K_{2}$, we iteratively call Algorithm \ref{alg:net_morph_general}
and shrink the size of $K_{2}^{r}$ until the loss converges to 0.
This iteration shall terminate as we are able to guarantee that if
$K_{2}^{r}=1$, the loss is 0. For the other case that $F_{l+1}$
expands $G$, the algorithm is similar.

\begin{figure}
\begin{centering}
\includegraphics[width=0.8\linewidth]{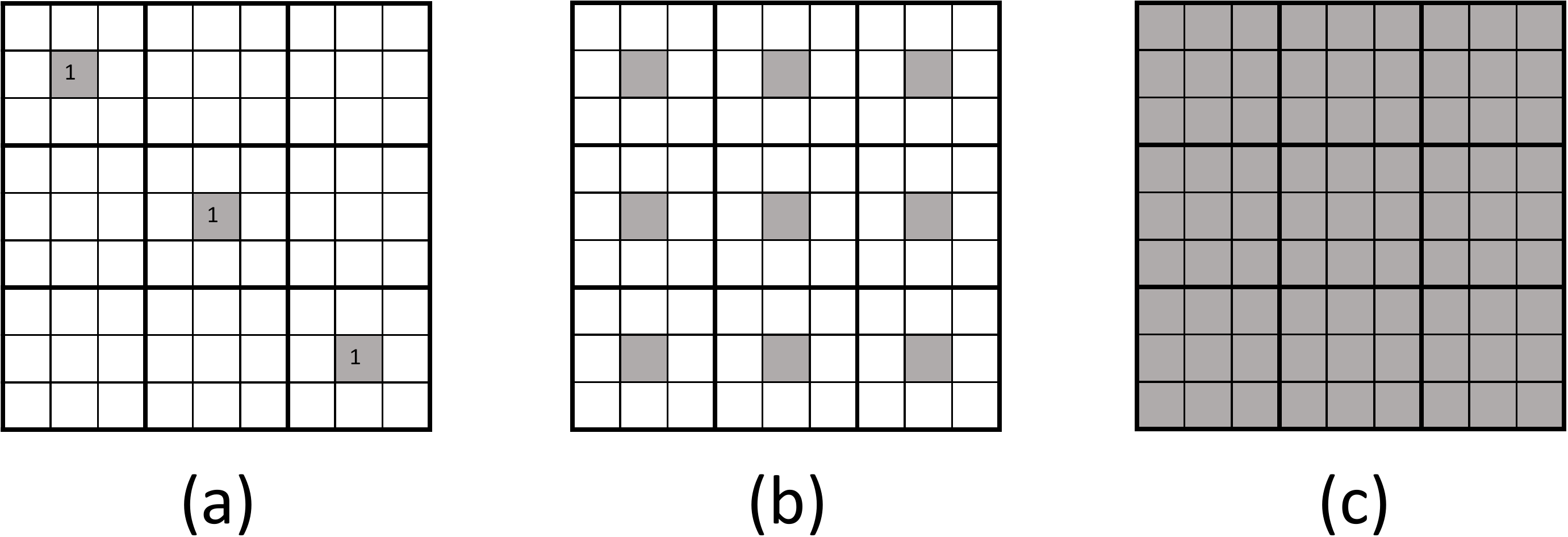}
\par\end{centering}

\protect\caption{Non-zero element (indicated as gray) occupations of different algorithms:
(a) IdMorph in $O(C)$, (b) NetMorph worst case in $O(C^{2})$, and
(c) NetMorph best case in $O(C^{2}K^{2})$. $C$ and $K$ represent
the channel size and kernel size. This figure shows a 4D convolutional
filter of shape $(3,3,3,3)$ flattened in 2D. It can be seen that
the filter in IdMorph is very sparse. \label{fig:net_morph_alg_comp}}

\end{figure}

The sacrifice of the non-sparse practice in Algorithm \ref{alg:net_morph_smart}
is illustrated in Fig. \ref{fig:net_morph_alg_comp}. In its worst
case, it might not be able to fill in all parameters with non-zero
elements, but still fill asymptotically. This figure compares the
non-zero element occupations for IdMorph and NetMorph. We assume $C_{l+1}=O(C_{l})\triangleq O(C)$.
In the best case (c), NetMorph is able to occupy all the elements
by non-zeros, with an order of $O(C^{2}K^{2})$. And in the worst
case (b), it has an order of $O(C^{2})$ non-zero elements. Generally,
NetMorph lies in between the best case and worst case. IdMorph (a)
only has an order of $O(C)$ non-zeros elements. Thus the non-zero
occupying rate of NetMorph is higher than IdMorph for at least one
order of magnitude. In practice, we shall also have $C\gg K$, and
thus NetMorph can asymptotically fill in all parameters with non-zero
elements.

\subsection{Network Morphism: Non-linear Case}

In the proposed network morphism it is also required to deal with
the non-linearities in a neural network. In general, it is not trivial
to replace the layer $B_{l+1}=\varphi(G\circledast B_{l+1})$ with
two layers $B_{l+1}=\varphi(F_{l+1}\circledast\varphi(F_{l}\circledast B_{l-1}))$,
where $\varphi$ represents the non-linear activation function.

For an idempotent activation function satisfying $\varphi\circ\varphi=\varphi,$
the IdMorph scheme in Net2Net \cite{ChenGoodfellowShlens2015} is
to set $F_{l+1}=I$, and $F_{l}=G$, where $I$ represents the identity
mapping. Then we have {\small{}
\begin{equation}
\varphi(I\circledast\varphi(G\circledast B_{l-1})=\varphi\circ\varphi(G\circledast B_{l+1})=\varphi(G\circledast B_{l+1}).
\end{equation}
} However, although IdMorph works for the ReLU activation function,
it cannot be applied to other commonly used activation functions,
such as Sigmoid and TanH, since the idempotent condition is not satisfied.

To handle arbitrary continuous non-linear activation functions, we
propose to define the concept of P(arametric)-activation function
family. A family of P-activation functions for an activation function
$\varphi$, can be defined to be any continuous function family that
maps $\varphi$ to the linear identity transform $\varphi_{id}:x\mapsto x$.
The P-activation function family for $\varphi$ might not be uniquely
defined. We define the canonical form for P-activation function family
as follows:
\begin{equation}
P\textit{-}\varphi\triangleq\{\varphi^{a}\}|_{a\in[0,1]}=\{(1-a)\cdot\varphi+a\cdot\varphi_{id}\}|_{a\in[0,1]},
\end{equation}
where $a$ is the parameter to control the shape morphing of the activation
function. We have $\varphi^{0}=\varphi$, and $\varphi^{1}=\varphi_{id}$.
The concept of P-activation function family extends PReLU \cite{HeZhangRenEtAl2015a},
and the definition of PReLU coincides with the canonical form of P-activation
function family for the ReLU non-linear activation unit.

\begin{figure}
\begin{centering}
\includegraphics[width=0.8\linewidth]{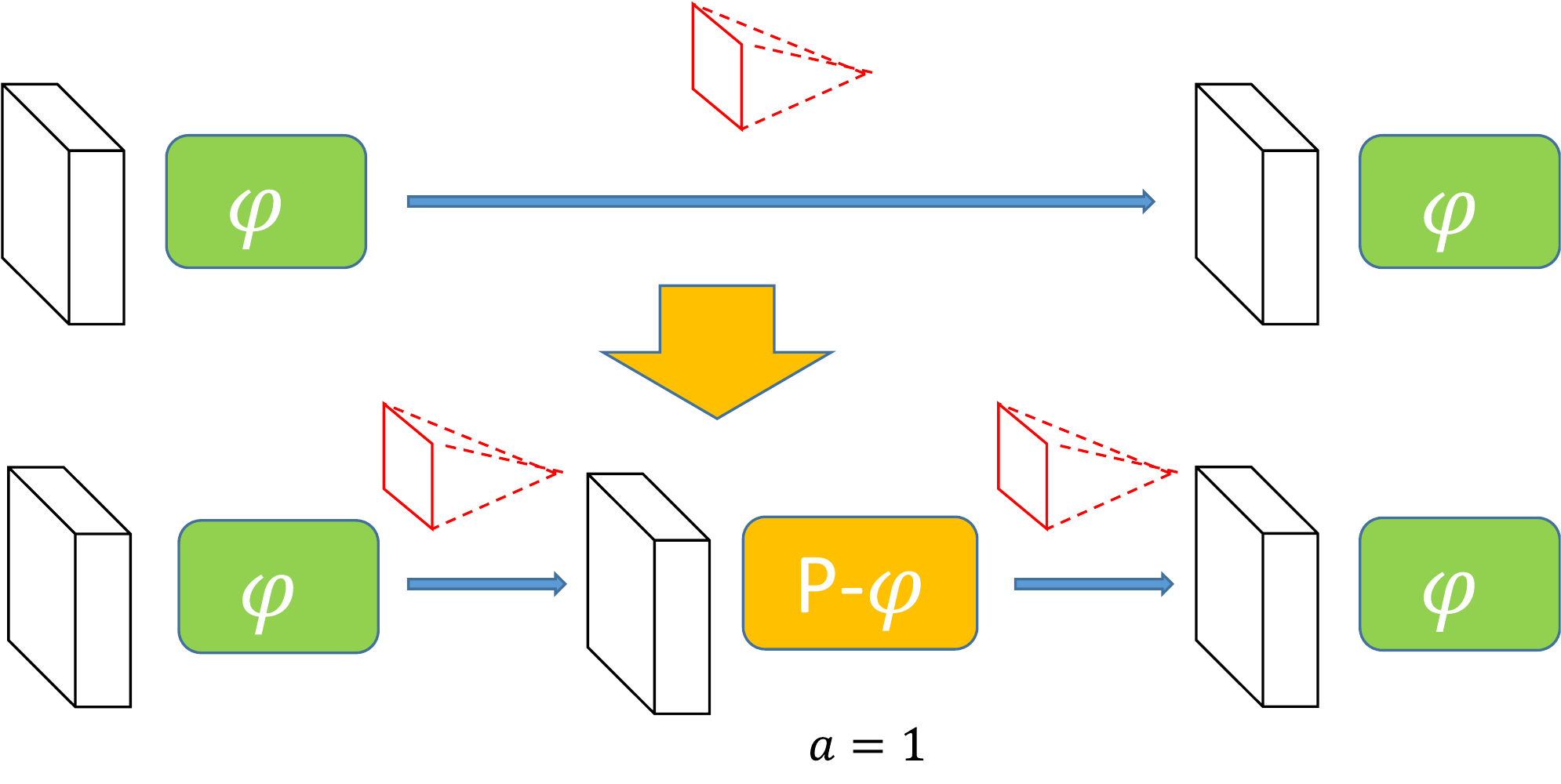}
\par\end{centering}

\protect\caption{Network morphism non-linear. Activations indicated as green can be
safely added; the activation in yellow needs to be set as linear ($a=1$)
at the beginning, and then is able to grow into a non-linear one as
$a$ is being learned. \label{fig:net_morph_nonlinear}}

\end{figure}

The idea of leveraging P-activation function family for network morphism
is shown in Fig. \ref{fig:net_morph_nonlinear}. As shown, it is safe
to add the non-linear activations indicated by the green boxes, but
we need to make sure that the yellow box is equivalent to a linear
activation initially. This linear activation shall grow into a non-linear
one once the value of $a$ has been learned. Formally, we need to
replace the layer $B_{l+1}=\varphi(G\circledast B_{l+1})$ with two
layers $B_{l+1}=\varphi(F_{l+1}\circledast\varphi^{a}(F_{l}\circledast B_{l-1}))$.
If we set $a=1$, the morphing shall be successful as long as the
network morphing equation (\ref{eq:net_morph}) is satisfied: {\small{}{}
\begin{align}
\varphi(F_{l+1}\circledast\varphi^{a}(F_{l}\circledast B_{l-1})) & =\varphi(F_{l+1}\circledast F_{l}\circledast B_{l-1})\\
 & =\varphi(G\circledast B_{l-1}).
\end{align}
}The value of $a$ shall be learned when we continue to train the
model.

\subsection{Stand-alone Width and Kernel Size Morphing}

\label{sub:netmorph_width}

As mentioned, the network morphism equation (\ref{eq:net_morph})
involves network depth, width, and kernel size morphing. Therefore,
we can conduct width and kernel size morphing by introducing an extra
depth morphing via Algorithm \ref{alg:net_morph_smart}.

Sometimes, we need to pay attention to stand-alone network width and
kernel size morphing operations. In this section, we introduce solutions
for these situations.

\subsubsection{Width Morphing}

For width morphing, we assume $B_{l-1},\ B_{l},\ B_{l+1}$ are all
parent network layers, and the target is to expand the width (channel
size) of $B_{l}$ from $C_{l}$ to $\tilde{C_{l}}$, $\tilde{C_{l}}\geq C_{l}$.
For the parent network, we have

{\small{}
\begin{equation}
B_{l}(c_{l})=\sum_{c_{l-1}}B_{l-1}(c_{l-1})*F_{l}(c_{l},c_{l-1}),
\end{equation}
\begin{equation}
B_{l+1}(c_{l+1})=\sum_{c_{l}}B_{l}(c_{l})*F_{l+1}(c_{l+1},c_{l}).
\end{equation}
} For the child network, $B_{l+1}$ should be kept unchanged: {\small{}{}
\begin{gather}
B_{l+1}(c_{l+1})=\sum_{\tilde{c_{l}}}B_{l}(\tilde{c_{l}})*\tilde{F}_{l+1}(c_{l+1},\tilde{c_{l}})\\
=\sum_{c_{l}}B_{l}(c_{l})*F_{l+1}(c_{l+1},c_{l})+\sum_{\bar{c_{l}}}B_{l}(\bar{c_{l}})*\tilde{F}_{l+1}(c_{l+1},\bar{c_{l}}),
\end{gather}
} where $\tilde{c_{l}}$ and $c_{l}$ are the indices of the channels
of the child network blob $\tilde{B_{l}}$ and parent network blob
$B_{l}$. $\bar{c_{l}}$ is the index of the complement $\tilde{c_{l}}\backslash c_{l}$.
Thus, we only need to satisfy: {\small{}{}
\begin{alignat}{1}
0 & =\sum_{\bar{c_{l}}}B_{l}(\bar{c_{l}})*\tilde{F}_{l+1}(c_{l+1},\bar{c_{l}})\\
 & =\sum_{\bar{c_{l}}}B_{l-1}(c_{l-1})*\tilde{F_{l}}(\bar{c_{l}},c_{l-1})*\tilde{F}_{l+1}(c_{l+1},\bar{c_{l}}),
\end{alignat}
} or simply,
\begin{equation}
\tilde{F_{l}}(\bar{c_{l}},c_{l-1})*\tilde{F}_{l+1}(c_{l+1},\bar{c_{l}})=0.
\end{equation}
It is obvious that we can either set $\tilde{F_{l}}(\bar{c_{l}},c_{l-1})$
or $\tilde{F}_{l+1}(c_{l+1},\bar{c_{l}})$ to 0, and the other can
be set arbitrarily. Following the non-sparse practice, we set the
one with less parameters to 0, and the other one to random noises.
The zeros and random noises in $\tilde{F_{l}}$ and $\tilde{F}_{l+1}$
may be clustered together. To break this unwanted behavior, we perform
a random permutation on $\tilde{c_{l}}$, which will not change $B_{l+1}$.


\subsubsection{Kernel Size Morphing}
\begin{figure}
\begin{centering}
\includegraphics[width=0.75\linewidth]{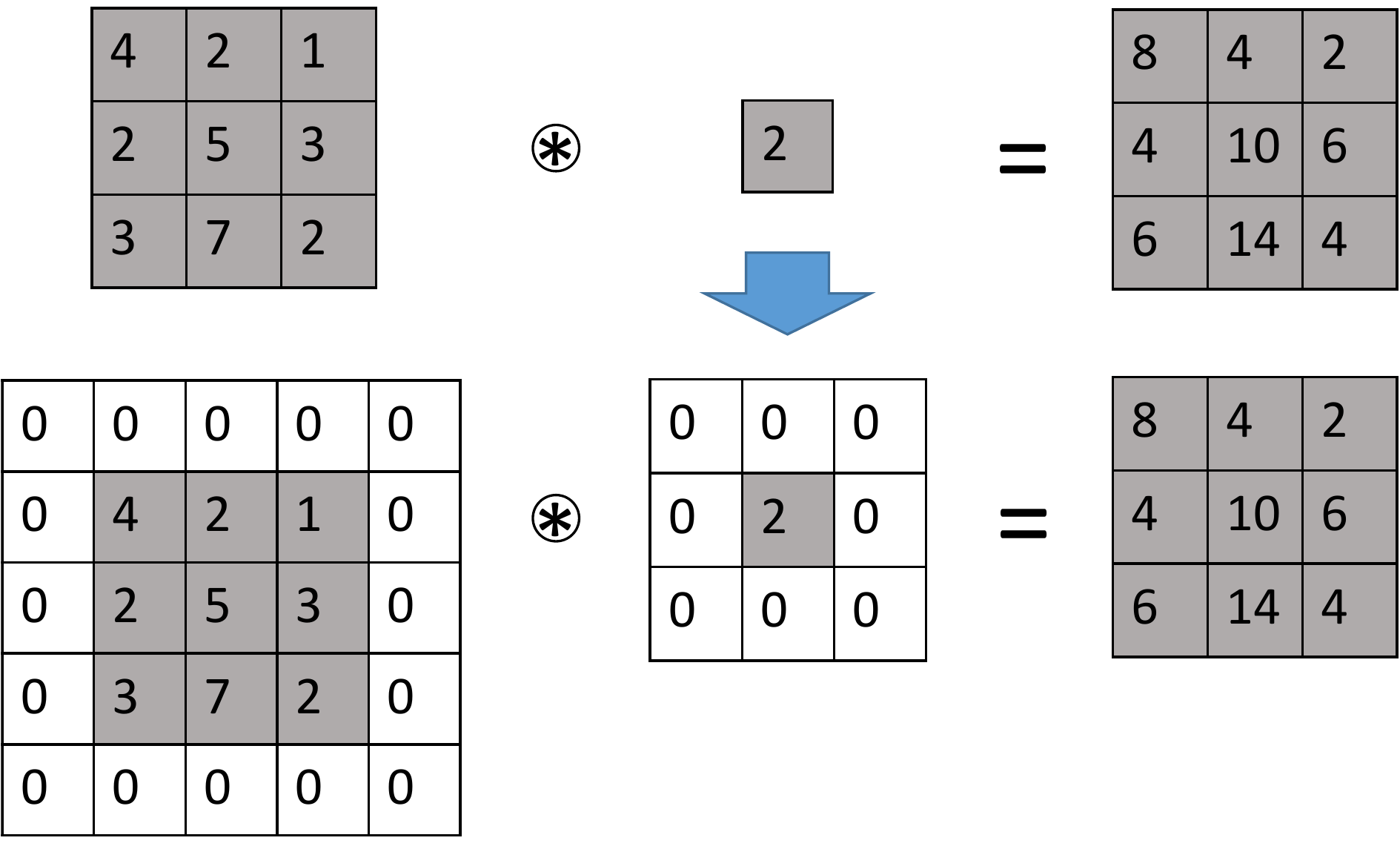}
\par\end{centering}
 \protect\caption{Network morphism in kernel size. Both the filters and blobs are padded
with the same size of zeros around to keep the final results unchanged.
\label{fig:net_morph_ksize}}
\end{figure}

For kernel size morphing, we propose a heuristic yet effective solution.
Suppose that a convolutional layer $l$ has kernel size of $K_{l}$,
and we want to expand it to $\tilde{K_{l}}$. When the filters of
layer $l$ are padded with $(\tilde{K_{l}}-K_{l})/2$ zeros on each
side, the same operation shall also apply for the blobs. As shown
in Fig. \ref{fig:net_morph_ksize}, the resulting blobs are of the
same shape and also with the same values.

\subsection{Subnet Morphing}

\begin{figure*}[t]
\begin{centering}
\subfloat[Typical subnets]{\protect

\begin{centering}
\protect\includegraphics[width=0.14\linewidth]{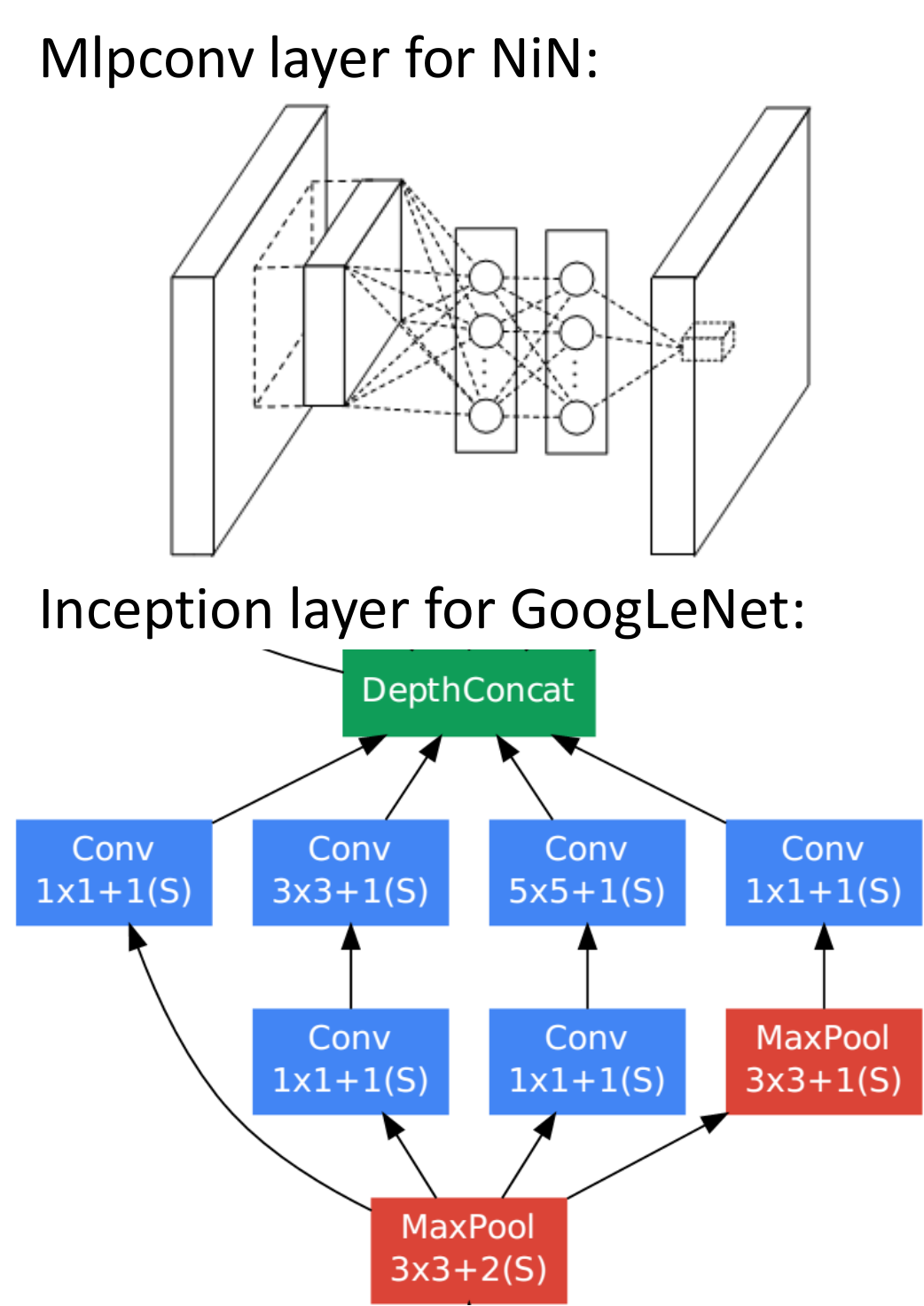}
\protect
\par\end{centering}

}\subfloat[Sequential subnet morphing]{\protect

\begin{centering}
\protect\includegraphics[width=0.3\linewidth]{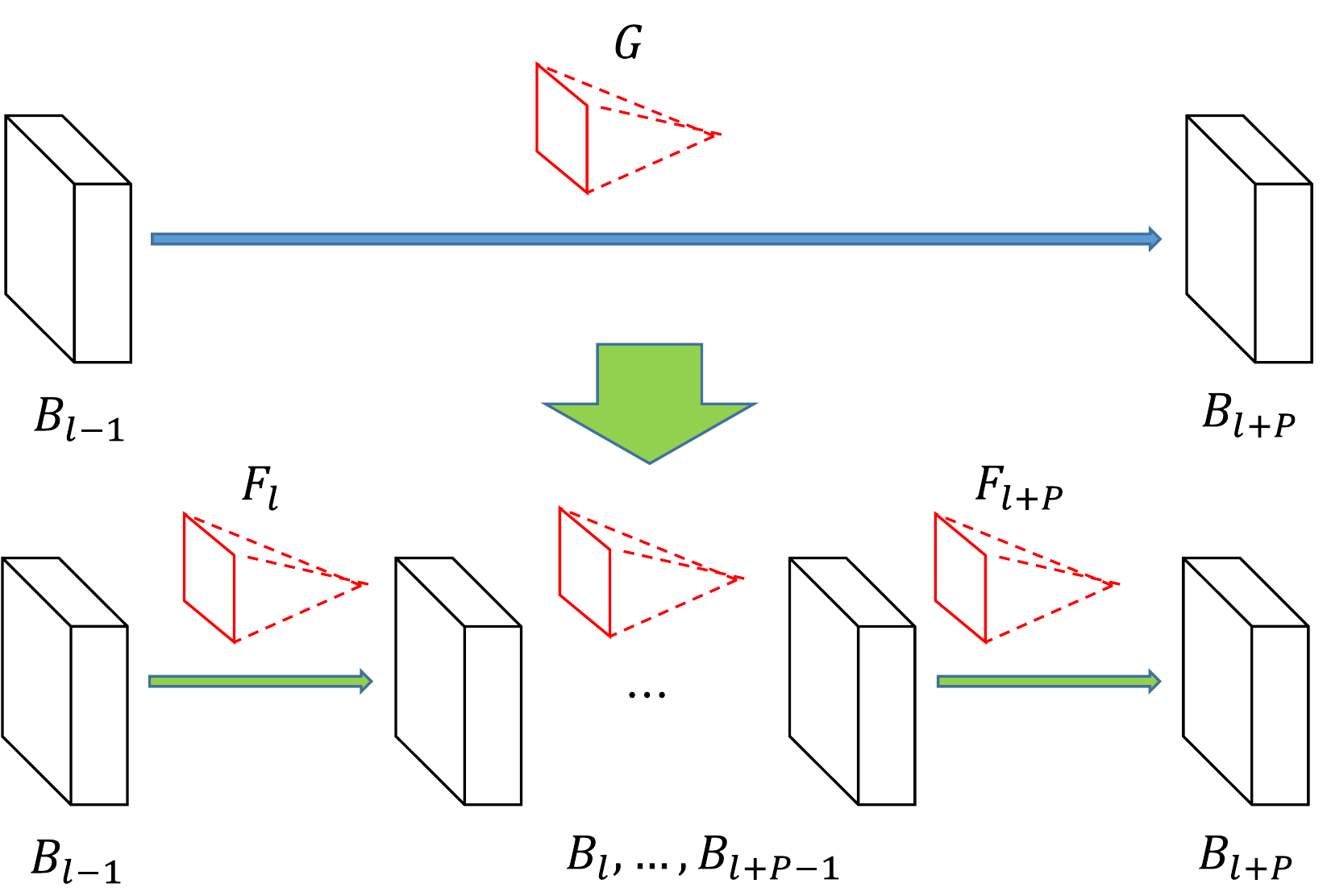}
\protect
\par\end{centering}

}\subfloat[Stacked sequential subnet morphing]{\protect

\begin{centering}
\protect\includegraphics[width=0.52\linewidth]{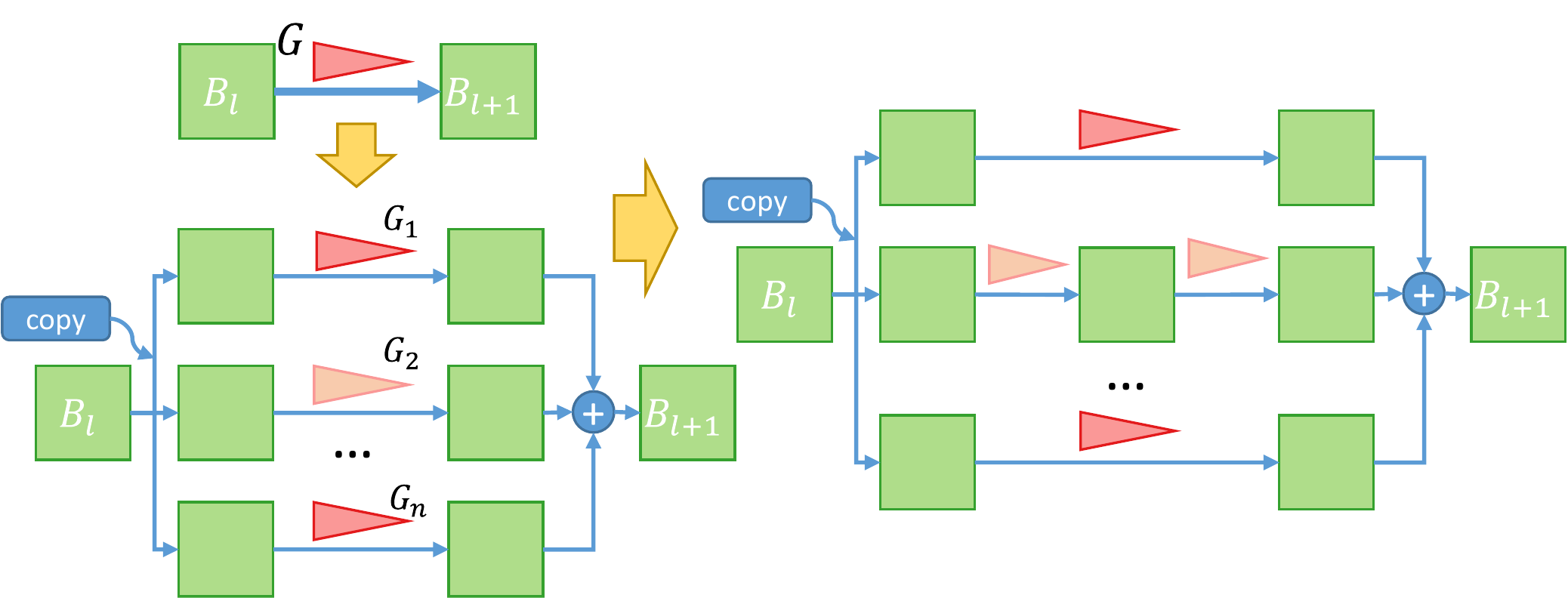}
\protect
\par\end{centering}

}
\par\end{centering}

 \protect\caption{Subnet morphing. (a) Subnet examples of the mlpconv layer in NiN and
inception layer in GoogleNet. (b) Sequential subnet morphing from
a single layer to $P+1$ layers. (c) Workflow for stacked sequential
subnet morphing. \label{fig:net_morph_subnet}}

\end{figure*}

Modern networks are going deeper and deeper. It is challenging to
manually design tens of or even hundreds of layers. One elegant strategy
is to first design a subnet template, and then construct the network
by these subnets. Two typical examples are the mlpconv layer for Network
in Network (NiN) \cite{LinChenYan2013} and the inception layer for
GoogLeNet \cite{SzegedyLiuJiaEtAl2014}, as shown in Fig. \ref{fig:net_morph_subnet}(a).

We study the problem of subnet morphing in this section, that is,
network morphism from a minimal number (typically one) of layers in
the parent network to a subnet in the child network. One commonly
used subnet is the stacked sequential subnet as shown in Fig. \ref{fig:net_morph_subnet}(c).
An exmaple is the inception layer for GoogLeNet with a four-way stacking
of sequential subnets.

We first describe the morphing operation for the sequential subnet,
based on which its stacked version is then obtained.

Sequential subnet morphing is to morph from a single layer to multiple
sequential layers, as illustrated in Fig. \ref{fig:net_morph_subnet}(b).
Similar to Equation (\ref{eq:net_morph}), one can derive the network
morphism equation for sequential subnets from a single layer to $P+1$
layers: {\small{}{}
\begin{equation}
\tilde{G}(c_{l+P},c_{l-1})=\sum_{\mathclap{c_{l},\cdots,c_{l+P-1}}}F_{l}(c_{l},c_{l-1})*\cdots*F_{l+P}(c_{l+P},c_{l+P-1}),\label{eq:net_morph_subnet}
\end{equation}
} where $\tilde{G}$ is a zero-padded version of $G$. Its effective
kernel size is $\tilde{K}=\sum_{p=0,\cdots,P}K_{l+p}-P$, and $K_{l}$
is the kernel size of layer $l$. Similar to Algorithm \ref{alg:net_morph_general},
subnet morphing equation (\ref{eq:net_morph_subnet}) can be solved
by iteratively optimizing the parameters for one layer with the parameters
for the other layers fixed. We can also develop a practical version
of the algorithm that can solve for Equation (\ref{eq:net_morph_subnet}),
which is similar to Algorithm \ref{alg:net_morph_smart}. The algorithm
details are omitted here.

For stacked sequential subnet morphing, we can follow the work flow
illustrated as Fig. \ref{fig:net_morph_subnet}(c). First, a single
layer in the parent network is split into multiple paths. The split
$\{G_{i}\}$ is set to satisfy
\begin{equation}
{\displaystyle \sum_{i=1}^{n}G_{i}=G},
\end{equation}
in which the simplest case is $G_{i}={\textstyle \frac{1}{n}}G$.
Then, for each path, a sequential subnet morphing can be conducted.
In Fig. \ref{fig:net_morph_subnet}(c), we illustrate an $n$-way
stacked sequential subnet morphing, with the second path morphed into
two layers.

\section{Experimental Results}

\label{sec:exp}

In this section, we conduct experiments on three datasets (MNIST,
CIFAR10, and ImageNet) to show the effectiveness of the proposed network
morphism scheme, on 1) different morphing operations, 2) both the
classic and convolutional neural networks, and 3) both the idempotent
activations (ReLU) and non-idempotent activations (TanH).

\subsection{Network Morphism for Classic Neural Networks}

\begin{figure}
\begin{centering}
\subfloat[PReLU]{\begin{centering}
\includegraphics[width=0.5\linewidth]{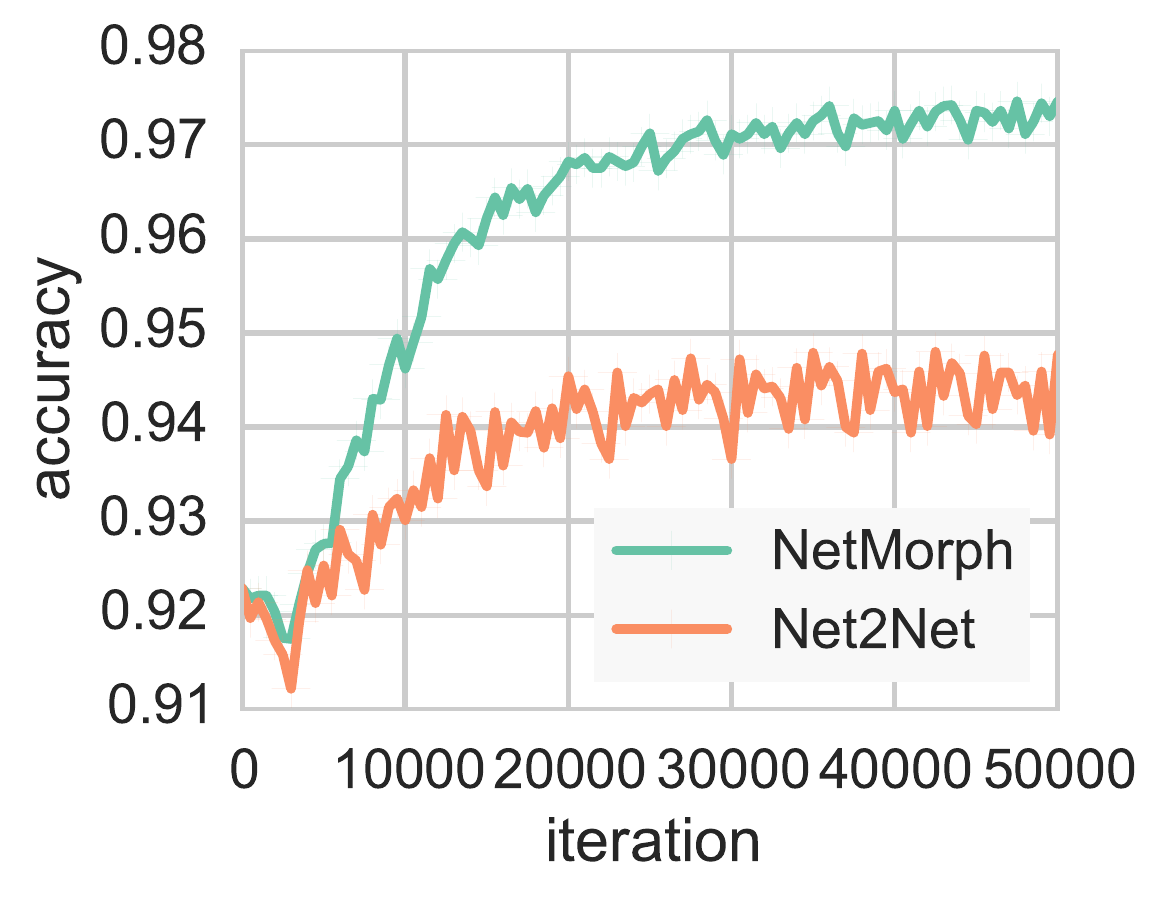}
\par\end{centering}

}\subfloat[PTanH]{\begin{centering}
\includegraphics[width=0.5\linewidth]{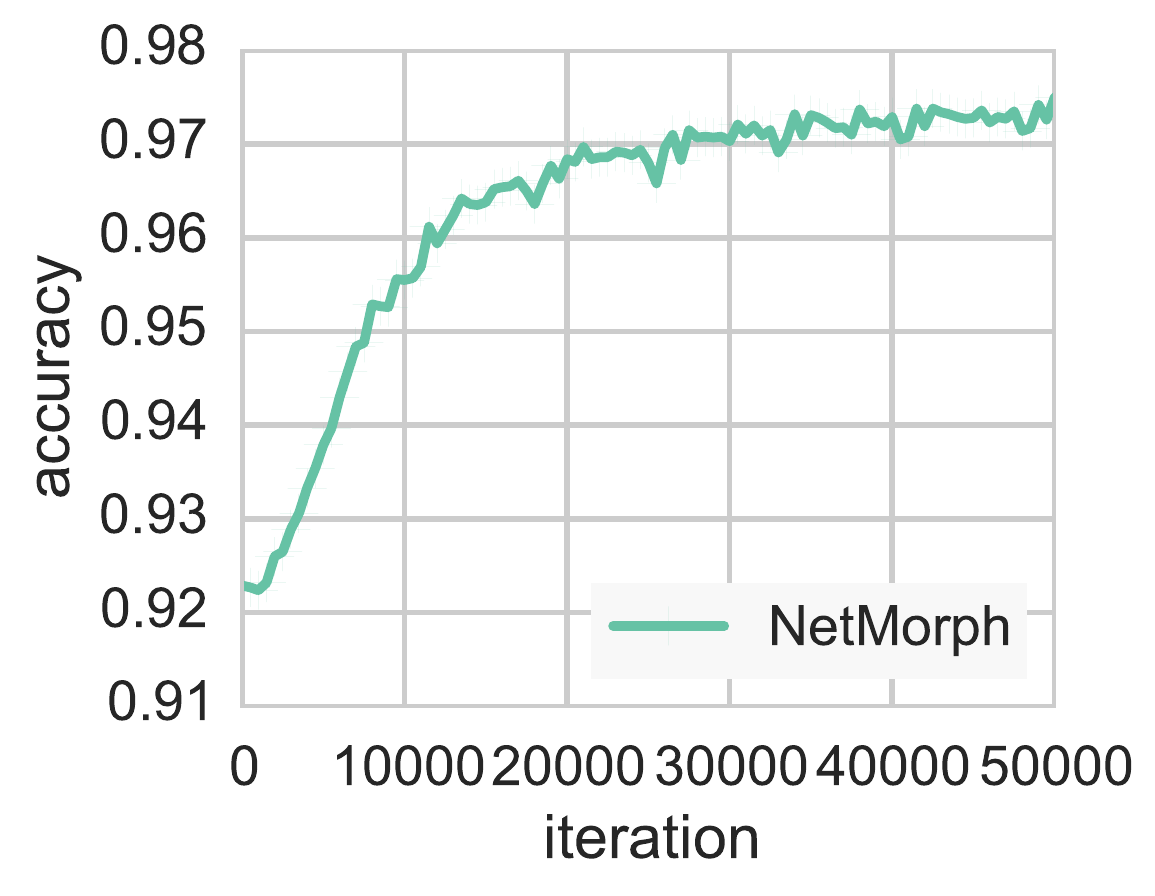}
\par\end{centering}

}
\par\end{centering}

\caption{Morphing on MNIST from softmax regression to multiple layer perception.
\label{fig:mnist_netmorph}}

\end{figure}

The first experiment is conducted on the MNIST dataset \cite{LeCunBottouBengioEtAl1998}.
MNIST is a standard dataset for handwritten digit recognition, with
60,000 training images and 10,000 testing images. In this section,
instead of using state-of-the-art DCNN solutions \cite{LeCunBottouBengioEtAl1998,ChangChen2015},
we adopt the simple softmax regression model as the parent network
to evaluate the effectiveness of network morphism on classic networks.
The gray-scale $28\times28$ digit images were flattened as a $784$
dimension feature vector as input. The parent model achieved 92.29\%
accuracy, which is considered as the baseline. Then, we morphed this
model into a multiple layer perception (MLP) model by adding a PReLU
or PTanH hidden layer with the number of hidden neurons $h=50$. Fig.~\ref{fig:mnist_netmorph}(a)
shows the performance curves of NetMorph\footnote{In the experiments, we use NetMorph to represent the proposed network
morphism algorithm.} 
and Net2Net after morphing. We can see that, for the PReLU activation,
NetMorph works much better than Net2Net. NetMorph continues to improve
the performance from 92\% to 97\%, while Net2Net improves only to
94\%. We also show the curve of NetMorph with the non-idempotent activation
PTanH in Fig. \ref{fig:mnist_netmorph}(b). The curve for Net2Net
is unavailable since it cannot handle non-idempotent activations.

\subsection{Depth Morphing, Subnet Morphing, and Internal Regularization for
DCNN}

\begin{figure*}
\begin{centering}
\subfloat[cifar\_111 $\to$ 211]{\begin{centering}
\includegraphics[width=0.25\linewidth]{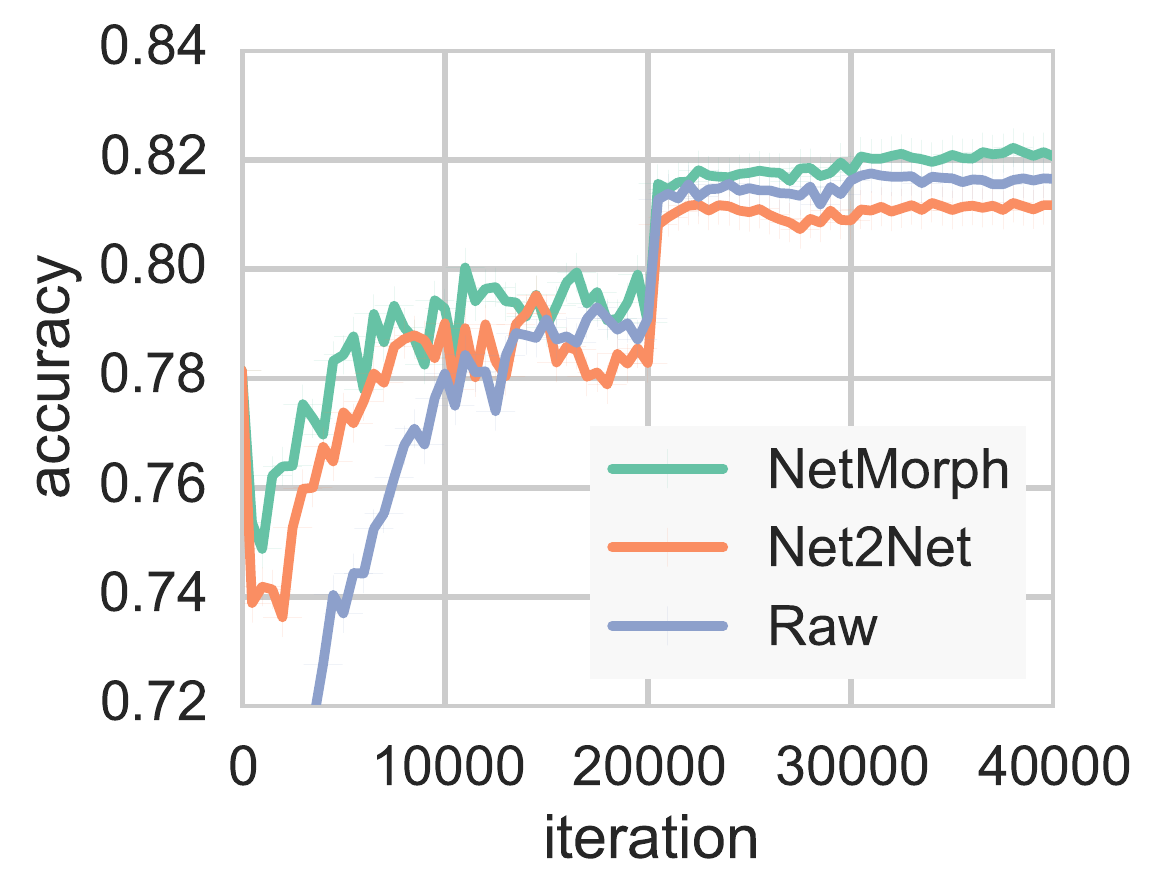}
\par\end{centering}

}\subfloat[cifar\_211 $\to$ 222]{\begin{centering}
\includegraphics[width=0.25\linewidth]{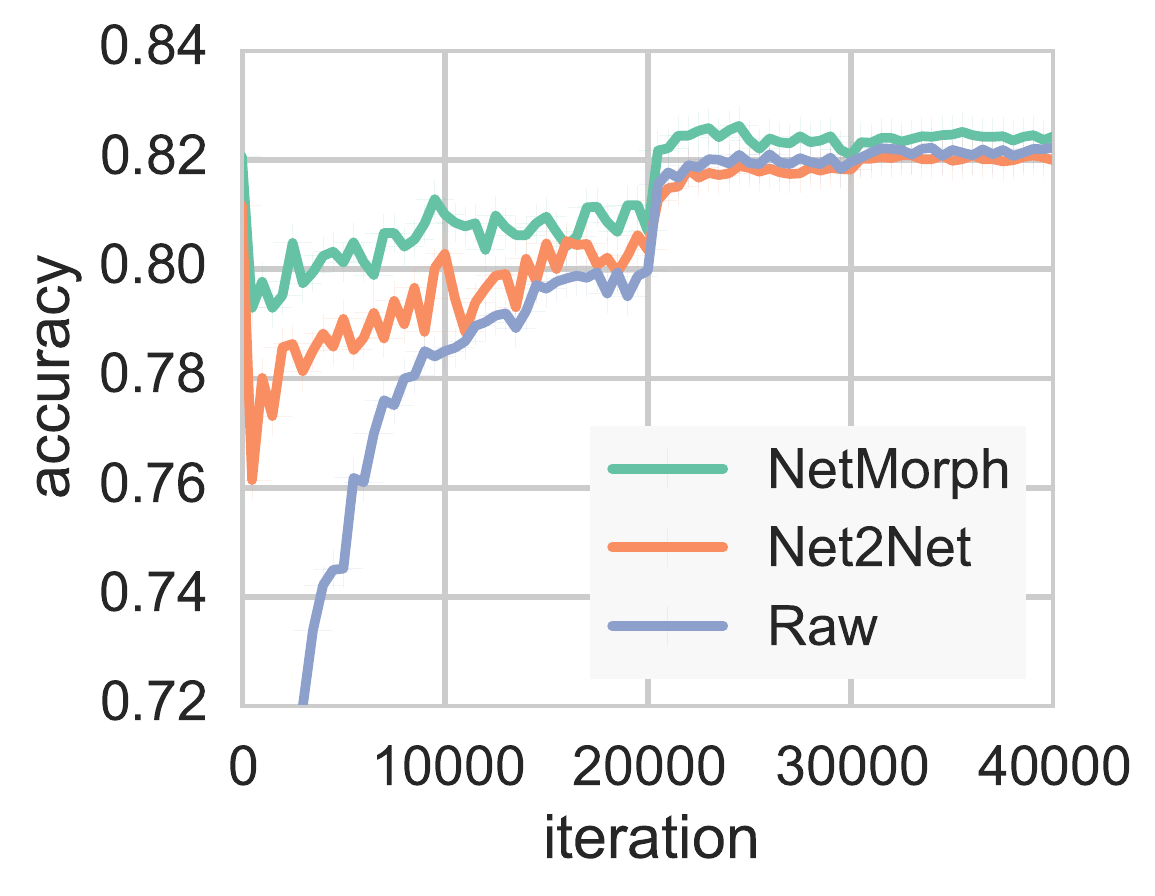}
\par\end{centering}

}\subfloat[cifar\_222 $\to$ 2222]{\begin{centering}
\includegraphics[width=0.25\linewidth]{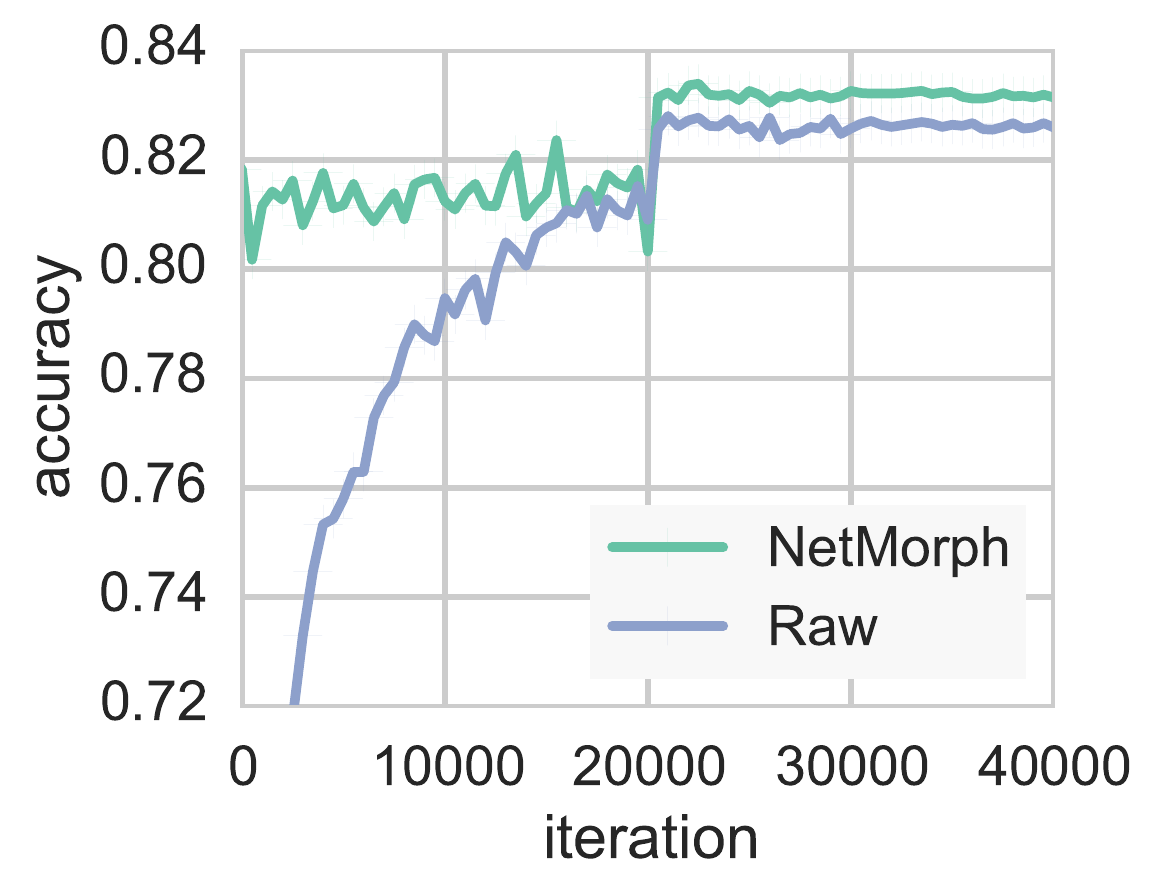}
\par\end{centering}

}\subfloat[cifar\_2222 $\to$ 3333]{\begin{centering}
\includegraphics[width=0.25\linewidth]{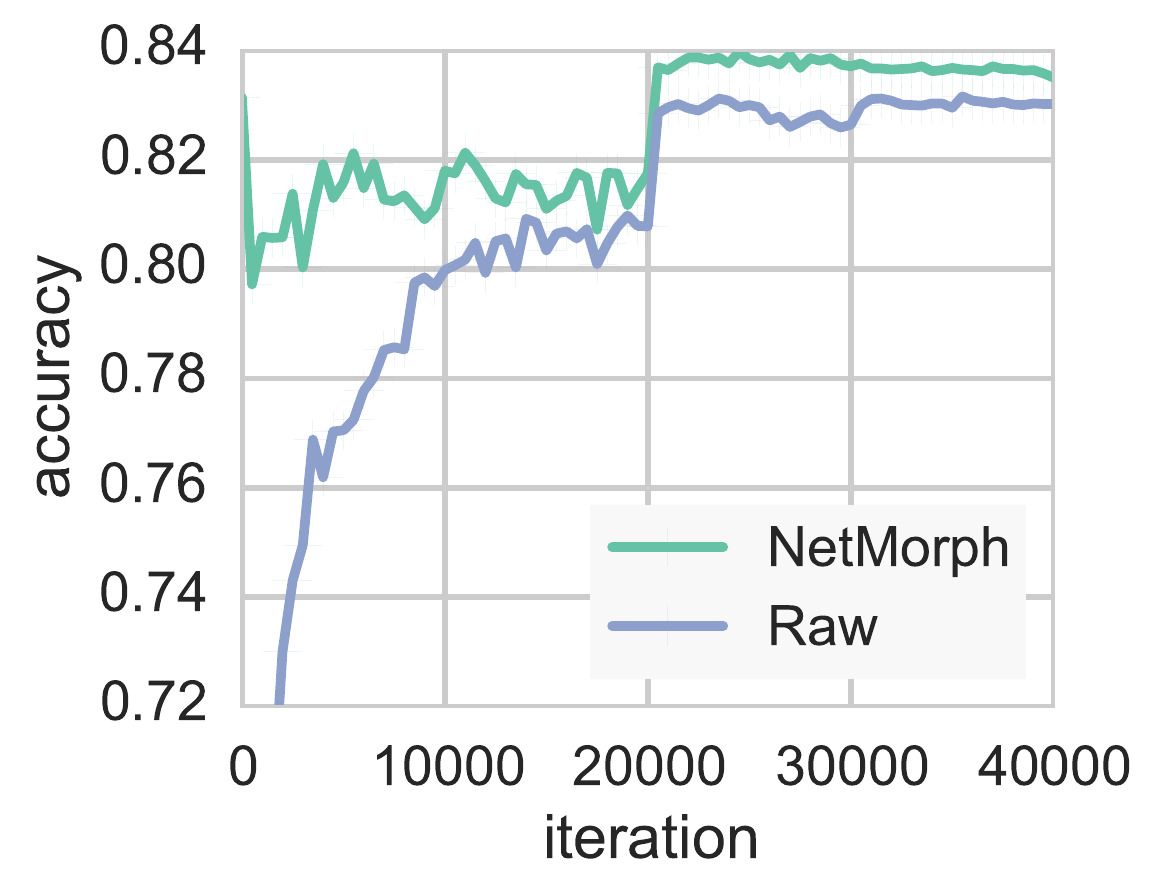}
\par\end{centering}

}
\par\end{centering}


\caption{Depth morphing and subnet morphing on CIFAR10.\label{fig:results_cifar10_depth}}

\end{figure*}

Extensive experiments were conducted on the CIFAR10 dataset \cite{KrizhevskyHinton2009}
to verify the network morphism scheme for convolutional neural networks.
CIFAR10 is an image recognition database composed of $32\times32$
color images. It contains 50,000 training images and 10,000 testing
images for ten object categories. The baseline network we adopted
is the Caffe \cite{JiaShelhamerDonahueEtAl2014} \texttt{cifar10\_quick}
model with an accuracy of 78.15\%.

In the following, we use the unified notation \texttt{cifar\_ddd}
to represent a network architecture of three subnets, in which each
digit \texttt{d} is the number of convolutional layers in the corresponding
subnet. The detailed architecture of each subnet is described by \texttt{(<kernel\_size>:<num\_output>,...)}.
Following this notation, the \texttt{cifar10\_quick} model, which
has three convolutional layers and two fully connected layers, can be denoted as \texttt{cifar\_111}, with its architecture described with \texttt{(5:32)(5:32)(5:64)}. Additionally, we also
use \texttt{{[}<subnet>{]}} to indicate the grouping of layers in
a subnet, and \texttt{x<times>} to represent for the repetition of
layers or subnets. Hence, \texttt{cifar10\_quick} can also be denoted
as \texttt{(5:32)x2(5:64)} or \texttt{{[}(5:32){]}x2{[}(5:64){]}}.
Note that in this notation, the fully connected layers are ignored.

Fig.~\ref{fig:results_cifar10_depth} shows the comparison results
between NetMorph and Net2Net, in the morphing sequence of \texttt{cifar\_111$\to$211$\to$222$\to$2222$\to$3333}.
The detailed network architectures of these networks are shown in
Table \ref{tab:net_arch_cifar}. In this table, some layers are morphed
by adding a 1x1 convolutional layer with channel size four times larger.
This is a good practice adopted in the design of current state-of-the-art
network \cite{HeZhangRenEtAl2015}. Algorithm \ref{alg:net_morph_smart}
is leveraged for the morphing. From Fig. 8(a) and (b), we can see
the superiority of NetMorph over Net2Net. NetMorph improves the performance
from 78.15\% to 82.06\%, then to 82.43\%, while Net2Net from 78.15\%
to 81.21\%, then to 81.99\%. The relatively inferior performance of
Net2Net may be caused by the IdMorph in Net2Net involving too many
zero elements on the embedded layer, while non-zero elements are also
not in a consistent scale with existing parameters. We also verified
this by comparing the histograms of the embedded filters (after morphing)
for both methods. The parameter scale for NetMorph fits a normal distribution
with a relatively large standard derivation, while that of Net2Net
shows two peaks around 0 and 0.5.

Fig.~\ref{fig:results_cifar10_depth}(c) illustrates the performance
of NetMorph for subnet morphing. The architecture is morphed from
\texttt{cifar\_222} to\texttt{ cifar\_2222}. As can be seen, NetMorph
achieves additional performance improvement from 82.43\% to 83.14\%.
Fig.~\ref{fig:results_cifar10_depth}(d) illustrates for the morphing
from \texttt{cifar\_2222 }to\texttt{ cifar\_3333}, and the performance
is further improved to around 84\%.

Finally, we compare NetMorph with the model directly trained from
scratch (denoted as Raw) in Fig. \ref{fig:results_cifar10_depth}.
It can be seen that NetMorph consistently achieves a better accuracy
than Raw. As the network goes deeper, the gap becomes larger. We interpret
this phenomena as the internal regularization ability of NetMorph.
In NetMorph, the parameters are learned in multiple phases rather
than all at once. Deep neural networks usually involve a large amount
of parameters, and overfit to the training data can occur easily.
For NetMorph, the parameters learned have been placed in a good position
in the parameter space. We only need to explore for a relatively small
region rather than the whole parameter space. Thus, the NetMorph learning
process shall result in a more regularized network to achieve better
performance.

\subsection{Kernel Size Morphing and Width Morphing}

\label{sub:exp_cifar10_width}

\begin{figure}
\begin{centering}
\subfloat[Kernel size morphing]{\begin{centering}
\includegraphics[width=0.5\linewidth]{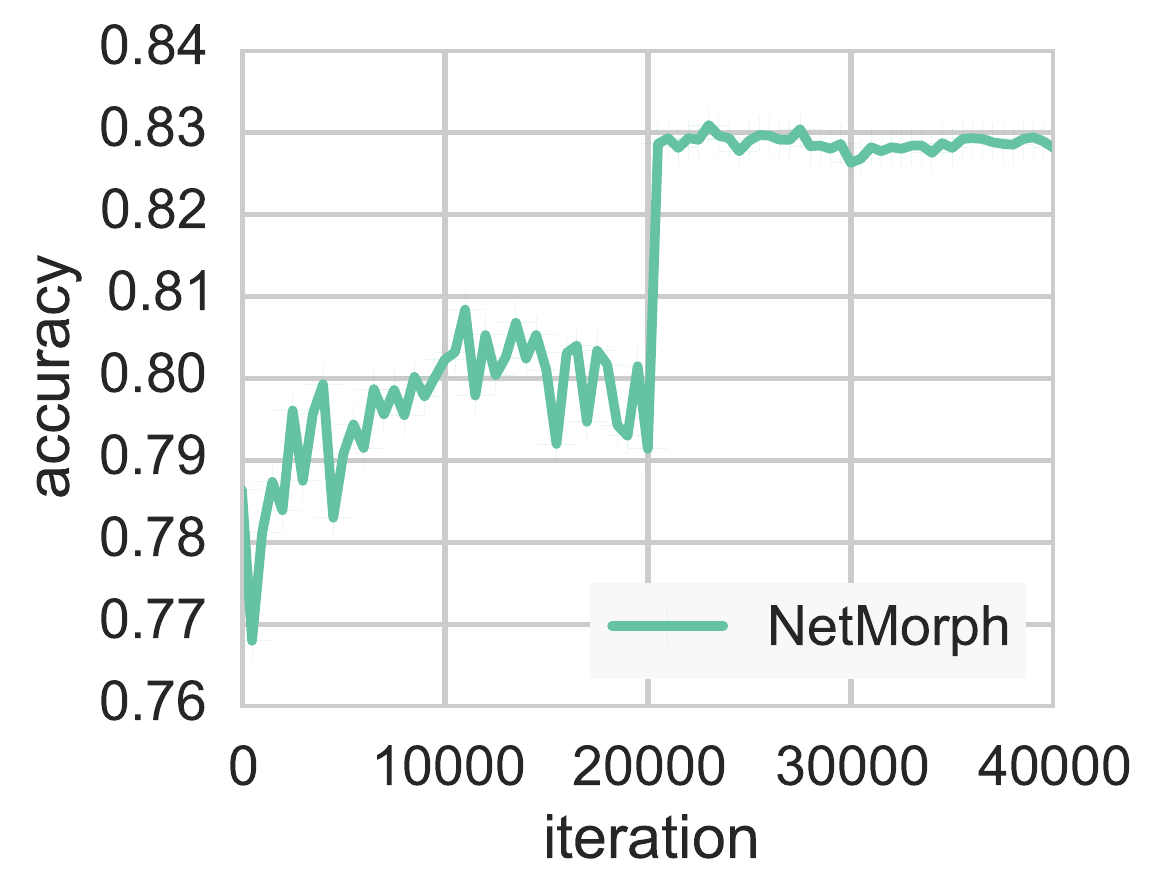}
\par\end{centering}

}\subfloat[Width morphing]{\begin{centering}
\includegraphics[width=0.5\linewidth]{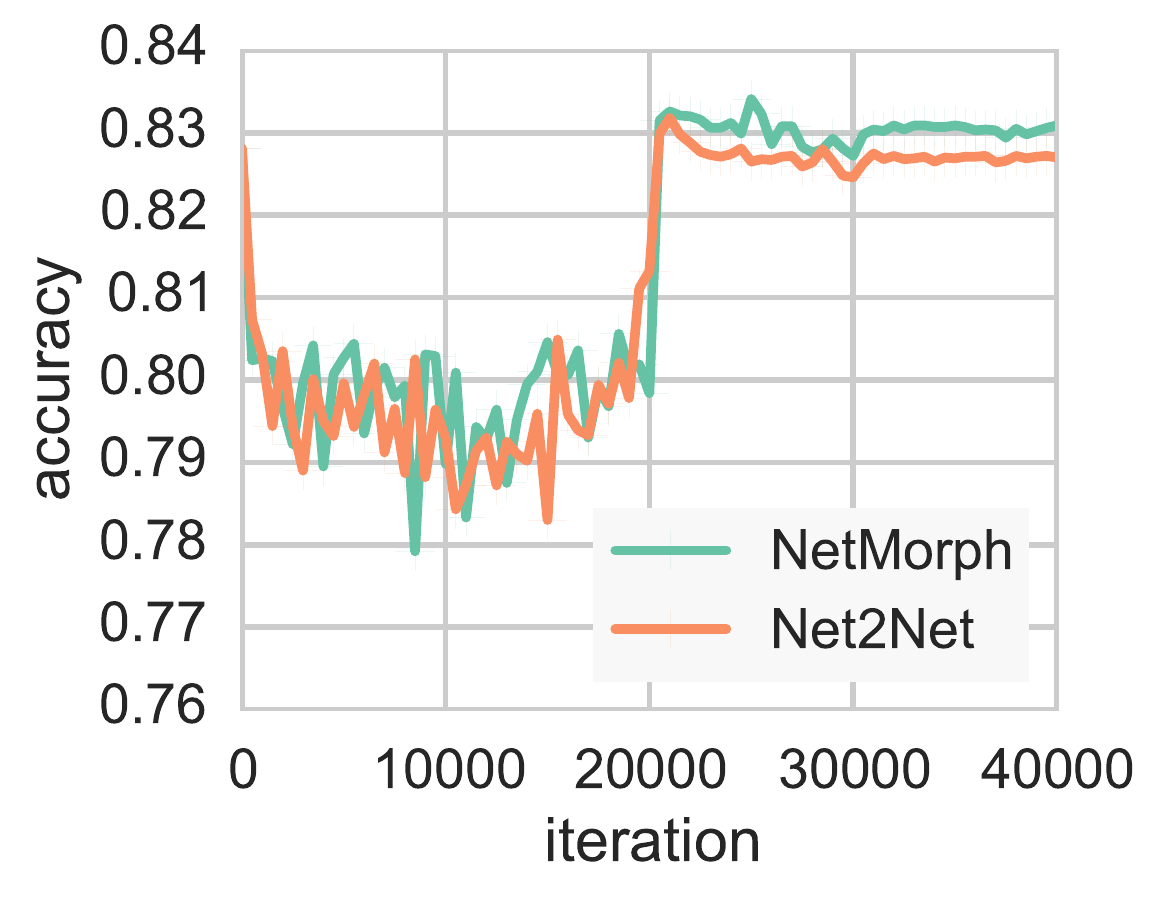}
\par\end{centering}

}
\par\end{centering}


\caption{Kernel size and width morphing on CIFAR10. \label{fig:results_cifar10_width}}

\end{figure}

\begin{table*}[t]
\begin{centering}
\caption{Network architectures for the experiments on CIFAR10. \label{tab:net_arch_cifar}}

\par\end{centering}

\begin{centering}

\par\end{centering}

\begin{centering}
\begin{tabular}{|c|c|}
\hline
Scheme & Network Architecture\tabularnewline
\hline
\hline
\texttt{cifar\_111} & \texttt{(5:32)(5:32)(5:64)}\tabularnewline
\hline
\texttt{cifar\_211} & \texttt{(5:32x4)(1:32)(5:32)(5:64)}\tabularnewline
\hline
\texttt{cifar\_222} & \texttt{{[}(5:32x4)(1:32){]}x2{[}(5:64x4)(1:64){]}}\tabularnewline
\hline
\texttt{cifar\_2222} & \texttt{{[}(5:32x4)(1:32){]}x2{[}(5:64x4)(1:64){]}x2}\tabularnewline
\hline
\texttt{cifar\_3333} & \texttt{{[}(5:32x4)(3:32x4)(1:32){]}x2{[}(5:64x4)(3:64x4)(1:64){]}x2}\tabularnewline
\hline
\hline
\texttt{cifar\_base} & \texttt{(5:32)(1:32)(5:32)(1:32)(5:64)(1:64)}\tabularnewline
\hline
\texttt{cifar\_ksize} & \texttt{(5:32)(3:32)(5:32)(3:32)(5:64)(3:64)}\tabularnewline
\hline
\texttt{cifar\_width} & \texttt{(5:64)(3:32)(5:64)(3:32)(5:128)(3:64)}\tabularnewline
\hline
\end{tabular}
\par\end{centering}

\centering{}
\end{table*}

\begin{table*}[t]
\begin{centering}
\caption{Network architectures for the experiments on ImageNet. \label{tab:net_arch_imagenet}}

\par\end{centering}

\begin{centering}

\par\end{centering}

\begin{centering}
\begin{tabular}{|c|c|}
\hline
Scheme & Network Architecture\tabularnewline
\hline
\hline
{VGG16} & \texttt{{[}(3:64)x2{]}{[}(3:128)x2{]}{[}(3:256)x3{]}{[}(3:512)x3{]}{[}(3:512)x3{]}}\tabularnewline
\hline
{VGG19} & \texttt{{[}(3:64)x2{]}{[}(3:128)x2{]}{[}(3:256)x4{]}{[}(3:512)x4{]}{[}(3:512)x4{]}}\tabularnewline
\hline
\multirow{2}{*}{{VGG16 (NetMorph)}} & \texttt{{[}(3:64)(3:64x4)(1:64){]}{[}(3:128)(3:128x4)(1:128){]}}\tabularnewline
 & \texttt{{[}(3:256)(3:256)(3:256x4)(1:256){]}{[}(3:512)x3{]}{[}(3:512)x3{]}}\tabularnewline
\hline
\end{tabular}
\par\end{centering}

\centering{}
\end{table*}

In this section we evaluate kernel size morphing and width morphing in the sequence
of \texttt{cifar\_base$\to$ksize$\to$width}. The detailed network architectures are shown in Table \ref{tab:net_arch_cifar}. The baseline
network \texttt{(cifar\_base)} is a narrower version of \texttt{cifar\_222}
with an accuracy of 81.48\%.

Fig. \ref{fig:results_cifar10_width}(a) shows the curve of kernel
size morphing, which expands the kernel size of the second layer in
each subnet from 1 to 3 (\texttt{cifar\_ksize}). This results in a
performance of 82.81\%, which is 1.33\% higher than the parent network.
We further double the number of channels (width) for the first layer
in each subnet (\texttt{cifar\_width)}. Fig. \ref{fig:results_cifar10_width}(b)
shows the results of NetMorph and Net2Net. As can be seen, NetMorph
is slightly better. It improves the performance to 83.09\% while Net2Net
dropped a little to 82.70\%.

For width morphing, NetMorph works for arbitrary continuous non-linear
activation functions, while Net2Net only for piece-wise linear ones.
We also conducted width morphing directly from the parent network
for TanH neurons, which results in about 4\% accuracy improvement.

\subsection{Experiment on ImageNet}

\label{sub:exp_imagenet}

We also conduct experiments on the ImageNet dataset \cite{RussakovskyDengSuEtAl2014}
with 1,000 object categories. The models were trained on 1.28 million
training images and tested on 50,000 validation images. The top-1
and top-5 accuracies for both 1-view and 10-view are reported.

The proposed experiments is based on the VGG16 net, which was actually
trained with multi-scales~\cite{SimonyanZisserman2014}. Because
the Caffe \cite{JiaShelhamerDonahueEtAl2014} implementation favors
single-scale, for a fair comparison, we first de-multiscale this model
by continuing to train it on the ImageNet dataset with the images
resized to $256\times256$. This process caused about 1\% performance
drop. This coincides with Table 3 in \cite{SimonyanZisserman2014}
for model D. In this paper, we adopt the de-multiscaled version of
the VGG16 net as the parent network to morph. The morphing operation
we adopt is to add a convolutional layer at the end of the first three
subsets for each. The detailed network architecture is shown in Table
\ref{tab:net_arch_imagenet}. We continue to train the child network
after morphing, and the final model is denoted as VGG16(NetMorph).
The results are shown in Table \ref{tab:results_imagenet}. We can
see that, VGG16(NetMorph) not only outperforms its parent network,
i.e, VGG16(baseline), but also outperforms the multi-scale version,
i.e, VGG16(multi-scale). Since VGG16(NetMorph) is a 19-layer network,
we also list the VGG19 net in Table \ref{tab:results_imagenet} for
comparison. As can be seen, VGG16(NetMorph) also outperforms VGG19
in a large margin. Note that VGG16(NetMorph) and VGG19 have different
architectures, as shown in Table \ref{tab:net_arch_imagenet}. Therefore,
the proposed NetMorph scheme not only can help improve the performance,
but also is an effective network architecture explorer. Further, we
are able to add more layers into VGG16(NetMorph), and a better performing
model shall be expected.

We compare the training time cost for the NetMorph learning scheme
and the training from scratch. VGG16 was trained for around 2\textasciitilde{}3
months for a single GPU time \cite{SimonyanZisserman2014}, which
does not include the pre-training time on a 11-layered network. For
a deeper network, the training time shall increase. While for the
19-layered network VGG16(NetMorph), the whole morphing and training
process was finished within 5 days, resulting in around 15x speedup.

\begin{table}
\caption{Comparison results on ImageNet. \label{tab:results_imagenet}}


\begin{centering}
\begin{footnotesize} %
\begin{tabular}{|c|c|c|c|c|}
\hline
 & Top-1  & Top-5  & Top-1  & Top-5\tabularnewline
 & 1-view  & 1-view  & 10-view  & 10-view\tabularnewline
\hline
{\scriptsize{}{}VGG16 (multi-scale)}  & 68.35\%  & 88.45\%  & 69.59\%  & 89.02\%\tabularnewline
\hline
{\scriptsize{}{}VGG19 (multi-scale)}  & 68.48\%  & 88.44\%  & 69.44\%  & 89.21\%\tabularnewline
\hline
\hline
{\scriptsize{}{}VGG16 (baseline)}  & 67.30\%  & 88.31\%  & 68.64\%  & 89.10\%\tabularnewline
\hline
{\scriptsize{}{}VGG16 (NetMorph)}  & \textbf{69.14\%}  & \textbf{89.00\%}  & \textbf{\noun{70.32\%}}  & \textbf{\noun{89.86\%}}\tabularnewline
\hline
\end{tabular}\end{footnotesize}
\par\end{centering}

\end{table}

\section{Conclusions}

In this paper, we have introduced the systematic study about network
morphism. The proposed scheme is able to morph a well-trained parent
network to a new child network, with the network function completely
preserved. The child network has the potential to grow into a more
powerful one in a short time. We introduced diverse morphing operations,
and developed novel morphing algorithms based on the morphism equations
we have derived. The non-linearity of a neural network has been carefully
addressed, and the proposed algorithms enable the morphing of any
continuous non-linear activation neurons. Extensive experiments have
been carried out to demonstrate the effectiveness of the proposed
network morphism scheme.

\bibliographystyle{icml2016}
\bibliography{dcnn}

\begin{thebibliography}{23}
\providecommand{\natexlab}[1]{#1}
\providecommand{\url}[1]{\texttt{#1}}
\expandafter\ifx\csname urlstyle\endcsname\relax
  \providecommand{\doi}[1]{doi: #1}\else
  \providecommand{\doi}{doi: \begingroup \urlstyle{rm}\Url}\fi

\bibitem[Ba \& Caruana(2014)Ba and Caruana]{BaCaruana2014}
Ba, Jimmy and Caruana, Rich.
\newblock Do deep nets really need to be deep?
\newblock In \emph{Advances in Neural Information Processing Systems}, pp.\
  2654--2662, 2014.

\bibitem[Bishop(2006)]{Bishop2006}
Bishop, Christopher~M.
\newblock Pattern recognition.
\newblock \emph{Machine Learning}, 2006.

\bibitem[Bucilu et~al.(2006)Bucilu, Caruana, and
  Niculescu-Mizil]{BuciluCaruanaNiculescu-Mizil2006}
Bucilu, Cristian, Caruana, Rich, and Niculescu-Mizil, Alexandru.
\newblock Model compression.
\newblock In \emph{Proceedings of the 12th ACM SIGKDD International Conference
  on Knowledge Discovery and Data Mining}, pp.\  535--541. ACM, 2006.

\bibitem[Chang \& Chen(2015)Chang and Chen]{ChangChen2015}
Chang, Jia-Ren and Chen, Yong-Sheng.
\newblock Batch-normalized maxout network in network.
\newblock \emph{arXiv preprint arXiv:1511.02583}, 2015.

\bibitem[Chen et~al.(2015)Chen, Goodfellow, and
  Shlens]{ChenGoodfellowShlens2015}
Chen, Tianqi, Goodfellow, Ian, and Shlens, Jonathon.
\newblock Net2net: Accelerating learning via knowledge transfer.
\newblock \emph{arXiv preprint arXiv:1511.05641}, 2015.

\bibitem[Girshick(2015)]{Girshick2015}
Girshick, Ross.
\newblock Fast r-cnn.
\newblock \emph{arXiv preprint arXiv:1504.08083}, 2015.

\bibitem[Girshick et~al.(2014)Girshick, Donahue, Darrell, and
  Malik]{GirshickDonahueDarrellEtAl2014}
Girshick, Ross, Donahue, Jeff, Darrell, Trevor, and Malik, Jagannath.
\newblock Rich feature hierarchies for accurate object detection and semantic
  segmentation.
\newblock In \emph{IEEE Conference on Computer Vision and Pattern Recognition},
  pp.\  580--587. IEEE, 2014.

\bibitem[Glorot \& Bengio(2010)Glorot and Bengio]{GlorotBengio2010}
Glorot, Xavier and Bengio, Yoshua.
\newblock Understanding the difficulty of training deep feedforward neural
  networks.
\newblock In \emph{International Conference on Artificial Intelligence and
  Statistics}, pp.\  249--256, 2010.

\bibitem[He et~al.(2015{\natexlab{a}})He, Zhang, Ren, and
  Sun]{HeZhangRenEtAl2015}
He, Kaiming, Zhang, Xiangyu, Ren, Shaoqing, and Sun, Jian.
\newblock Deep residual learning for image recognition.
\newblock \emph{arXiv preprint arXiv:1512.03385}, 2015{\natexlab{a}}.

\bibitem[He et~al.(2015{\natexlab{b}})He, Zhang, Ren, and
  Sun]{HeZhangRenEtAl2015a}
He, Kaiming, Zhang, Xiangyu, Ren, Shaoqing, and Sun, Jian.
\newblock Delving deep into rectifiers: Surpassing human-level performance on
  imagenet classification.
\newblock \emph{arXiv preprint arXiv:1502.01852}, 2015{\natexlab{b}}.

\bibitem[Jia et~al.(2014)Jia, Shelhamer, Donahue, Karayev, Long, Girshick,
  Guadarrama, and Darrell]{JiaShelhamerDonahueEtAl2014}
Jia, Yangqing, Shelhamer, Evan, Donahue, Jeff, Karayev, Sergey, Long, Jonathan,
  Girshick, Ross, Guadarrama, Sergio, and Darrell, Trevor.
\newblock Caffe: Convolutional architecture for fast feature embedding.
\newblock In \emph{Proceedings of the ACM International Conference on
  Multimedia}, pp.\  675--678. ACM, 2014.

\bibitem[Krizhevsky \& Hinton(2009)Krizhevsky and Hinton]{KrizhevskyHinton2009}
Krizhevsky, Alex and Hinton, Geoffrey.
\newblock Learning multiple layers of features from tiny images, 2009.

\bibitem[Krizhevsky et~al.(2012)Krizhevsky, Sutskever, and
  Hinton]{KrizhevskySutskeverHinton2012}
Krizhevsky, Alex, Sutskever, Ilya, and Hinton, Geoffrey~E.
\newblock Imagenet classification with deep convolutional neural networks.
\newblock In \emph{Advances in Neural Information Processing Systems}, pp.\
  1097--1105, 2012.

\bibitem[LeCun et~al.(1998)LeCun, Bottou, Bengio, and
  Haffner]{LeCunBottouBengioEtAl1998}
LeCun, Yann, Bottou, L{\'e}on, Bengio, Yoshua, and Haffner, Patrick.
\newblock Gradient-based learning applied to document recognition.
\newblock \emph{Proceedings of the IEEE}, 86\penalty0 (11):\penalty0
  2278--2324, 1998.

\bibitem[Lin et~al.(2013)Lin, Chen, and Yan]{LinChenYan2013}
Lin, Min, Chen, Qiang, and Yan, Shuicheng.
\newblock Network in network.
\newblock \emph{CoRR}, abs/1312.4400, 2013.
\newblock URL \url{http://arxiv.org/abs/1312.4400}.

\bibitem[Long et~al.(2014)Long, Shelhamer, and
  Darrell]{LongShelhamerDarrell2014}
Long, Jonathan, Shelhamer, Evan, and Darrell, Trevor.
\newblock Fully convolutional networks for semantic segmentation.
\newblock \emph{arXiv preprint arXiv:1411.4038}, 2014.

\bibitem[Oquab et~al.(2014)Oquab, Bottou, Laptev, and
  Sivic]{OquabBottouLaptevEtAl2014}
Oquab, Maxime, Bottou, Leon, Laptev, Ivan, and Sivic, Josef.
\newblock Learning and transferring mid-level image representations using
  convolutional neural networks.
\newblock In \emph{IEEE Conference on Computer Vision and Pattern Recognition},
  pp.\  1717--1724. IEEE, 2014.

\bibitem[Ren et~al.(2015)Ren, He, Girshick, and Sun]{RenHeGirshickEtAl2015}
Ren, Shaoqing, He, Kaiming, Girshick, Ross, and Sun, Jian.
\newblock Faster r-cnn: Towards real-time object detection with region proposal
  networks.
\newblock In \emph{Advances in Neural Information Processing Systems}, pp.\
  91--99, 2015.

\bibitem[Romero et~al.(2014)Romero, Ballas, Kahou, Chassang, Gatta, and
  Bengio]{RomeroBallasKahouEtAl2014}
Romero, Adriana, Ballas, Nicolas, Kahou, Samira~Ebrahimi, Chassang, Antoine,
  Gatta, Carlo, and Bengio, Yoshua.
\newblock Fitnets: Hints for thin deep nets.
\newblock \emph{arXiv preprint arXiv:1412.6550}, 2014.

\bibitem[Russakovsky et~al.(2014)Russakovsky, Deng, Su, Krause, Satheesh, Ma,
  Huang, Karpathy, Khosla, Bernstein, et~al.]{RussakovskyDengSuEtAl2014}
Russakovsky, Olga, Deng, Jia, Su, Hao, Krause, Jonathan, Satheesh, Sanjeev, Ma,
  Sean, Huang, Zhiheng, Karpathy, Andrej, Khosla, Aditya, Bernstein, Michael,
  et~al.
\newblock Imagenet large scale visual recognition challenge.
\newblock \emph{International Journal of Computer Vision}, pp.\  1--42, 2014.

\bibitem[Simonyan \& Zisserman(2014)Simonyan and
  Zisserman]{SimonyanZisserman2014}
Simonyan, Karen and Zisserman, Andrew.
\newblock Very deep convolutional networks for large-scale image recognition.
\newblock \emph{arXiv preprint arXiv:1409.1556}, 2014.

\bibitem[Szegedy et~al.(2014)Szegedy, Liu, Jia, Sermanet, Reed, Anguelov,
  Erhan, Vanhoucke, and Rabinovich]{SzegedyLiuJiaEtAl2014}
Szegedy, Christian, Liu, Wei, Jia, Yangqing, Sermanet, Pierre, Reed, Scott,
  Anguelov, Dragomir, Erhan, Dumitru, Vanhoucke, Vincent, and Rabinovich,
  Andrew.
\newblock Going deeper with convolutions.
\newblock \emph{arXiv preprint arXiv:1409.4842}, 2014.

\bibitem[Weisstein(2002)]{Weisstein2002}
Weisstein, Eric~W.
\newblock \emph{CRC concise encyclopedia of mathematics}.
\newblock CRC press, 2002.

\end{thebibliography}

\end{document}